\documentclass{article}
\usepackage[utf8]{inputenc}
\usepackage{authblk}
\usepackage{setspace}
\usepackage[margin=1.25in]{geometry}
\usepackage{graphicx}
\graphicspath{ {./figures/} }
\usepackage{subcaption}
\usepackage{amsmath}
\usepackage{float}
\usepackage{xcolor}
\usepackage{booktabs} 
\usepackage{array} 
\usepackage{caption}
\usepackage{hyperref}
\usepackage{soul,xcolor}
\usepackage{mdframed}
\definecolor{lightyellow}{RGB}{255,255,153}
\usepackage{lmodern}
\usepackage{amssymb}
\usepackage{float}
\usepackage{placeins}
\usepackage{graphicx}
\usepackage{subcaption}
\usepackage{multirow}
\usepackage{hyperref}
\usepackage{xr-hyper}

\newcounter{suppnote}

\usepackage[backend=biber, style=nature, 
citestyle=numeric-comp,
sorting=none]{biblatex}
\title{Agronomic Characterization and Time-Series UAV Phenotyping to Support Field Pea Breeding in Iowa}
\author{Yanben Shen\textsuperscript{a}, Joscif Raigne\textsuperscript{a}, Asheesh Singh\textsuperscript{a}, Arti Singh\textsuperscript{a}*\\\textsuperscript{a}Department of Agronomy, Iowa state univeristy, Ames, U.S.A\\*Corresponding author: Arti singh

Email: arti@iastate.edu}

\begin{document}

\maketitle

\begin{abstract}
Field pea has gained interest in the Midwest U.S. and has the potential to benefit Iowa cropping systems. This study evaluated 17 accessions and 7 Canadian varieties for agronomic and seed yield performance. This study found that PI accession trials exhibit genetic diversity in phenology and canopy traits, whereas the Canadian elite varieties demonstrated great yield potential under Iowa conditions. The accession genotypes matured between 75 DAP (late June) and 95 DAP (late July) across six site-years. The top-yielding Canadian varieties averaged 1961 kg/ha across two site-years. To comprehensively assess field pea performance, this study extracted time-series Unmanned Aerial Vehicle(UAV)-derived traits and assessed UAV-derived traits and evaluated their correlations with multiple agronomic traits and yield. The results found that UAV-derived time-series canopy area was the most yield-relevant trait (r = 0.63) among all UAV-derived canopy features, vegetation indices, and manually collected measurements. Compared with yield, UAV-derived canopy area showed higher selection reliability (H² = 0.87, CV = 13.46\%), which can be used for phenotypic selection and can accelerate Midwest field pea breeding programs. This study identifies useful UAV-derived features for pea breeding and demonstrates the value of integrating diverse germplasm, elite commercial varieties, and time-series UAV phenotyping. Combining genetic variation from the accession trial with the yield potential of high-performing commercial varieties provides valuable breeding materials and phenotyping strategies for developing locally adapted, high-yielding field pea varieties for future production in the Midwest.

\end{abstract}
\newpage
\section{Introduction}

Field pea (\textit{Pisum sativum} L.) is a cool-season pulse crop valued for high seed protein content (21\% – 25\%) \cite{windsor2024characterizing,ren2021functionality}, nitrogen-fixing ability \cite{tulbek2024pea}, and a good candidate for diversifying and developing more sustainable cropping systems \cite{zhang2025integrated}. The U.S. is the fourth-largest field pea-producing country \cite{carr2024dryland}, and most of field pea production is in northern states, led by North Dakota, Montana, Washington state, and Idaho \cite{davis2025vegetables}. Field pea has gained more interest in the Midwest in the last two decades. Nebraska produced 24,040 metric tons in 2025, representing a 40\% increase in production from 2024 \cite{davis2025vegetables}. Integrating field peas as a replacement for chemical summer fallow not only improves soil fertility, long-term moisture retention, and nitrogen mineralization but also facilitates a highly productive pea rotational system. Together, pea-wheat rotation increases long-term sustainability and gains economic revenue in the dry land and semi-arid high plains regions \cite{koeshall2022replacing,nielsen2015cover}. Beyond Nebraska, interest in field peas has grown across the Central Great Plains. For instance, South Dakota State University Extension has conducted annual field pea variety trials since 2003, establishing a long-term benchmark that helps producers select well-adapted varieties \cite{SDSUExtension2025FieldPeaVarietyTrial}. Similarly, Kansas State University also conducted pea performance tests from 2014 to 2017 in the Upper Western part of Kansas. However, South Dakota, Nebraska, and Western Kansas differ from Iowa in climate, particularly because these regions generally receive less summer rainfall \cite{frankson2022nebraska,frankson2022iowa,frankson2022southdakota,frankson2022kansas}. Field pea has the potential to benefit Iowa cropping systems.

Early spring-planted peas followed by soybean or sorghum provided opportunities for crop rotation and improved land-use efficiency \cite{Fawcett2006DoubleCroppedFieldPea}. The total yield of double cropping of pea plus sorghum surpassed the soybean monoculture \cite{Fawcett2006DoubleCroppedFieldPea}. Regarding the yield impact of crop sequencing, research indicates that following after corn or soybean, pea yielded 96\% and 104\%, respectively \cite{KirkNDcropRotationPlanning}. Furthermore, corn and sorghum can be mutually beneficial. Alternating between sorghum and corn has been reported to increase yields; notably, corn following sorghum yielded over 18\% higher than continuous corn and achieved a similar performance to a traditional soybean-corn rotation \cite{UnitedSorghumCheckoffNDcroppingSystems}. Based on these findings, it's possible that a three-year rotation plan of soybean, double-cropping pea plus sorghum, and corn shows clear benefits for maximizing total crop productivity and regional land use. Despite the potential, there is no information on field pea yield performance tests available on field pea adaptation in Iowa. Important field pea information, agronomic performance, field management and operations, is lacking, making it difficult for researchers and farmers to this crop.   

High-throughput phenotyping is widely used to evaluate crop performance in field peas. Image analysis on the Unmanned Aerial Vehicle (UAV) images has been utilized for vegetation index \cite{quiros2019phenotyping}, aboveground biomass \cite{quiros2019phenotyping}, disease scoring \cite{zhao2022combining}, flowering detection and flowering time \cite{oppliger2025investigating}, canopy and plant height \cite{shen2022optimizing}, lodging \cite{bazrafkan2024predicting}, maturity \cite{bazrafkan2023predicting}, and yield prediction \cite{zhao2022combining}. In addition, time-series UAV phenotyping enables the capture of dynamic changes in canopy development throughout the growing season, including plant height \cite{alladassi2026persistent} and canopy color \cite{hassan2018time}, while also identifying critical plant features that contribute to final yield \cite{li2023time}. It provides more informative indicators of crop growth than single-date measurements, especially under variable field environments.

This study integrated time-series UAV phenotyping with conventional agronomic trait evaluation to comprehensively assess field pea performance in Iowa and identify genotypes with desirable breeding traits. Specifically, we evaluated relationships between UAV-derived and agronomic traits, identified yield-relevant canopy features, and compared time-series phenotyping with single-time-point measurements. This method can improve the efficiency of field pea evaluation and support breeding efforts for improved adaptation and yield potential in Midwestern environments.

\section{Method}

\subsection{Experimental design and management}
The field trial conducted in this study took place in 2022, 2024, and 2025 at Agricultural Engineering and Agronomy (AEA) Research Farms (42°02' latitude, -93° 77'longitude), Bruner (42°01', -93° 73'), and Burkey (42°01', -93° 78') research farm in Boone, Iowa, for a total of six site-years.  The experimental treatments were arranged in a randomized complete block design with two replications for each site-year.

In 2021, 14 accessions were selected from the USDA GRIN pea core collection \cite{coyne2005usda} based on days to maturity, yield, and hundred-seed weight. Along with these 14 accessions, three commercial checks (Delta, DS Admiral, and Mono 4010) from Albert Lea Seed (\textit{Albert Lea, MN}). These three varieties are unprotected varieties. Among these 17 genotypes, DS Admiral and Delta are semi-leafless, whereas the remaining 14 have normal leaf morphology. The semi-leafless type has greater lodging tolerance, lower disease severity, and higher yield than the normal leaf type \cite{shen2022evaluation}.

Seven Canadian lines were introduced to our lab through a material transfer agreement (MTA) and grown in three site-years immediately adjacent to the yield trial. The purpose of this experiment was to quantify the regional yield gap and set target yield baselines for the Midwest.

In the 2024 Bruner, 2024 Burkey, and 2025 Bruner environments, a new yield testing experiment, named as CDC set, was conducted using seven Canadian commercial field pea Varieties from the Crop Development Centre (CDC; Saskatoon, Canada): CDC Amarillo \cite{Amarillo}, CDC Canary \cite{Canary}, CDC Forest \cite{Forest}, CDC Hickie \cite{warkentincdcHickie}, CDC Lewochko \cite{Lewochko}, CDC Rider \cite{Rider}, and CDC Tollefson \cite{Tollefson}. These are elite varieties having very good stand ability and medium-long vine length. They are improved for powdery mildew resistance and are rated moderately resistant to Fusarium root rot. They have good seed characteristics and resistance to seed coat breakage, dimpling, and greenness \cite{pulse_variety_hub_2026}.

These varieties were planted in a separate block immediately adjacent to the trial containing the original 17 genotypes. All entries were planted on the same date, managed under the same field conditions, and evaluated using identical agronomic measurement protocols. The results of this side study are used to evaluate elite pea varieties in the Iowa environment.

The target seeding rate was 300,000 seeds acre$^{-1}$ in all years. In 2022, plots were planted using the drill planter and consisted of four rows, each 4.5 m long, with 25-cm row spacing. However, the planter was broken and could not be fixed. In 2024 and 2025, a new Seed Research Equipment Solutions Planter was purchased and used (\textit{Kincaid Equipment Manufacturing, Haven, KS, USA}). Alternatively, plots consisted of seven rows, each 4.5 m long, with 18.75-cm row spacing. Despite these differences in plot configuration, the target seeding rate per unit area remained consistent across years.

Field pea was planted in fields where corn had been grown the prior year. The heavy straw and chaff interfered with seed placement, reduced seed-to-soil contact, and caused uneven pea germination. Tillage was used with a chisel plow in the post-harvest period in the corn field, and a field cultivator was used in the pre-planting period to chop down the corn residue.  For soil fertility, fertilizer was applied in accordance with the annual soil report. Since nitrogen fertilizer was applied during the previous year's corn planting, they were not applied during pea planting.

Pre-planting herbicides, Dual Magnum (\textit{S-metolachlor}) and Prowl H2O (\textit{Pendimethalin}), were applied within 7 days before planting and incorporated into the soil. After pea emergence, post-herbicide, Select Max (\textit{Clethodim}) along with Pursuit (\textit{Imazethapyr}) were applied to control grass and broadleaf weeds. Plant counts were conducted after seedling emergence. Peas in a meter-square sampling area in the two center rows were counted at the three-leaf stage. The germination rate of each variety was over 75\% and there was no significant difference among treatments. Hand weeding was operated based on weed severity. 

Table \ref{tab:phenotyping_dates} details the planting and agronomic measurement dates for each site-year. Planting at the 2024 Bruner site was delayed by severe early-season waterlogging. Data for days to maturity were not collected in 2022 due to the application of glyphosate for crop desiccation. Measurements for early seedling vigor were introduced in 2024 as a new trait parameter.

\begin{table}[htbp]
\centering
\caption{Dates of planting and data collection for six site-years in central Iowa, U.S.}
\label{tab:phenotyping_dates}
\resizebox{\textwidth}{!}{
\begin{tabular}{lcccccc}
\hline 
Trait & 2022 AEA & 2022 Burkey & 2024 Bruner & 2024 Burkey & 2025 Bruner & 2025 Burkey \\ 

Planting date & 4/19/2022 & 4/19/2022 & 6/3/2024 & 4/10/2024 & 4/17/2025 & 4/17/2025 \\ 
\hline
Plant count & 6/1/2022 & 6/1/2022 & 6/20/2024 & 5/3/2024 & 5/12/2025 & 5/14/2025 \\ 
Early seedling vigor & -- & -- & 6/20/2024 & 5/10/2024 & 5/12/2025 & 5/14/2025 \\ 
Days to first flower & 5/29--6/16/2022 & 5/29--6/13/2022 & 6/27--7/20/2024 & 5/26--6/10/2024 & 5/26--6/16/2025 & 5/26--6/16/2025 \\ 
Canopy height & 6/28/2022 & 6/28/2022 & 7/31/2024 & 6/27/2024 & 6/30/2025 & 6/30/2025 \\ 
Plant height & 6/28/2022 & 6/28/2022 & 7/31/2024 & 6/27/2024 & 6/30/2025 & 6/30/2025 \\ 
Days to maturity & -- & -- & 8/12--9/4/2024 & 6/20--7/26/2024 & 6/16--7/19/2025 & 6/29--7/23/2025 \\ 
\hline
\end{tabular}  
}              
\end{table}

\subsection{Data collection}
Early seedling vigor was evaluated at 17-30 days after planting. The timing difference is due to plant growth varying across sites and years, and the trait was assessed after phenotypic variation appeared among genotypes. Early seedling vigor refers to the evaluation of plant emergence condition within the plot and the development of leaf area at the early stage \cite{brown2022utility,van2025unmanned}. Early seedling vigor incorporated multiple plant characteristics, including plant biomass, height, area, and leaf color, which are relevant to crop-weed competitiveness\cite{bertholdsson2005early}. Visual rating was evaluated in front of the plot and in the central rows using a 1-5 scale, with 1 representing a small, weak seedling and poor emergence, and 5 representing a large, vigorous seedling and full emergence \cite{van2025unmanned}. Mono 4010 variety is a forage variety having excellent seedling establishment and was used as a check to calibrate the visual rating standard \cite{han2012productivity}.

Days to flowering (DTF) were recorded when at least five plants had fully opened flowers. After the first genotype flowered, the field was evaluated every two days until the last genotype flowered. Days to maturity (DTM) were recorded when the pods had turned tan to brown, the seeds were loose within the pods, and the lower stems had dried and lost structural strength. Once the first genotype reached maturity, the field was evaluated every two days until the final genotype matured.

Canopy and plant height were measured at the pod-filling growth stage \cite{knott1987key}. Five random plants were selected from each plot. The distance from the soil surface to the end of the uppermost plant node was measured and considered to be the plant length. Canopy height measurements were taken simultaneously by measuring the canopy height of five randomly selected plants in the plot \cite{stelling1994performance}. 

The plants were harvested after being recorded as mature. The harvested plant material was dried at room temperature for 48 hours. The seeds were collected and cleaned using a BT-14 Single Plant Belt Thresher (\textit{ALMACO, Nevada, USA}) then weighed to determine yield.

\subsection{High throughput UAV phenotyping}
\begin{figure}[t]
    \centering
    \includegraphics[width=0.8\linewidth]{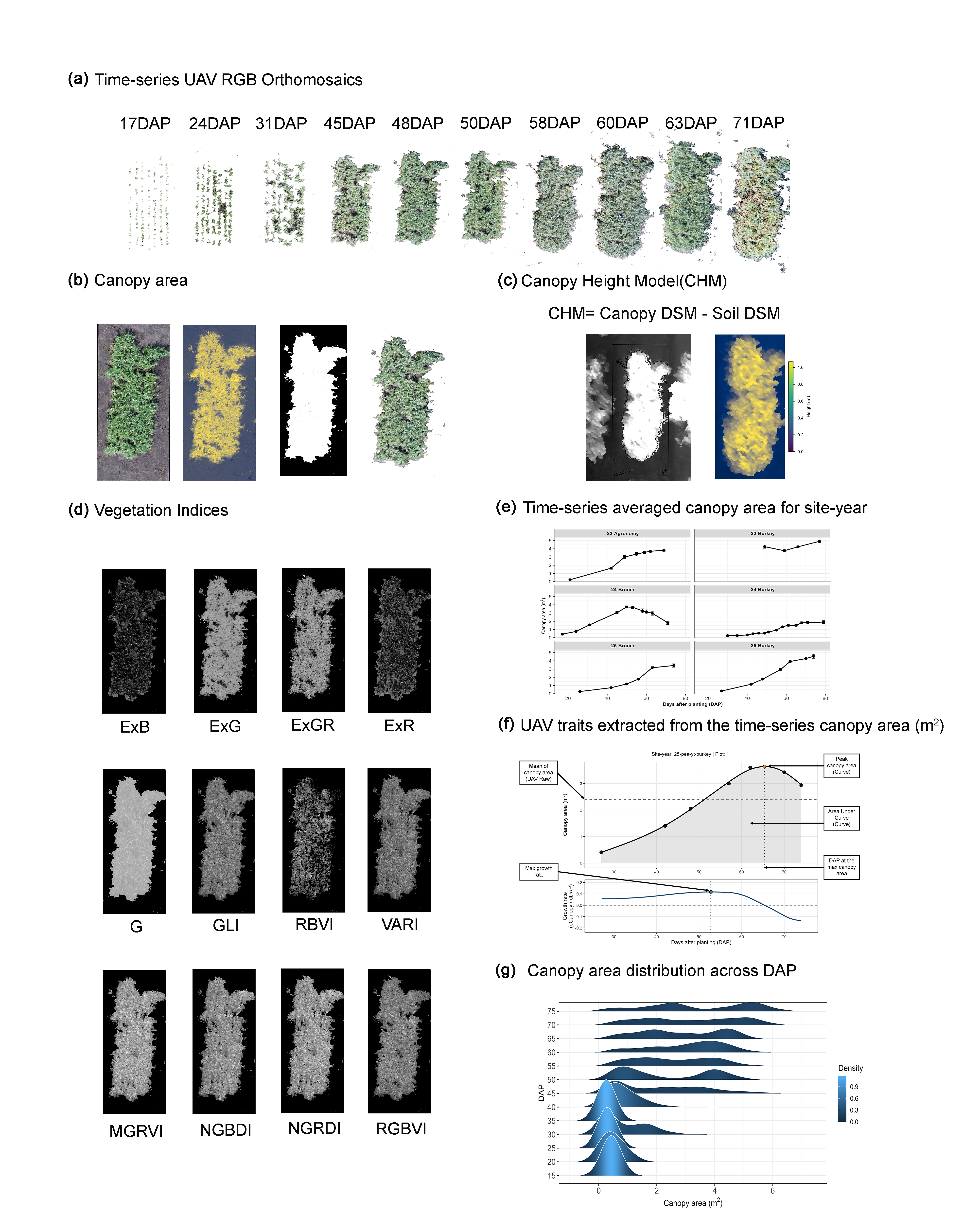}
    \caption{Overview of UAV-derived variable extraction from time-series RGB imagery. a)Example plot extracted from time-series RGB orthomosaics across seedling, early vegetative, late vegetative, reproductive, and maturing growth stages. b) Canopy area extraction and soil background removal using the Excess Green (ExG) vegetation index (threshold $> 0.1$). c) Canopy height model (CHM) calculated from canopy and soil digital surface models (DSMs). d) Extraction of RGB-based vegetation indices. The completed vegetation index list is in Supplementary Table S1. e) Time-series averaged canopy area for each site-year. Each dot represents a UAV flight date at a specific DAP, and the line shows the seasonal trend of canopy area within each site-year. f) UAV traits extracted from the GAM fitting curve in the time-series canopy area.  g) Time-series distribution of canopy area across DAP. The ridgeline density plots show changes in canopy area distribution across the growing season for all genotypes.}
    \label{fig:central_figure}
\end{figure}

UAV phenotyping began on the same day as seedling vigor data collection. Before flying, Four Group Control Points (GCPs) were installed at each field corner and one in the center. The geolocation of each center of GCP was recorded by an Emlid Reach RS RTK sensor. The Inspire 2 platform with a Zenmuse X5S RGB camera (DJI, Shenzhen, China) was used to take photos at 15 meters above ground level, with 80\% front to side image overlap. The image was stitched using Pix4D software (Pix4D S.A. Switzerland), which matched the GCPs and calibrated the overlaps to create a stitched orthomosaic map and a 2D-view digital elevation map (DEM) using the point clouds generated from the structure from motion photogrammetry. The Ground Sampling Distance (GSD) for DEM and orthomosaic images was 0.2 cm/pixel. UAV-derived canopy area, canopy height, and vegetation indices were extracted from the DEM and orthomosaic maps. Figure \ref{fig:central_figure} shows the plot-level canopy area, canopy height, and vegetation indices and temporal canopy traits throughout the time-series UAV images.

12 vegetation indices relevant to plant biomass, growth, leaf water content, and chlorophyll content \cite{woebbecke1995,meyer2008,gitelson2002,bannari1995,bendig2015,louhaichi2001,kataoka2003,bannari1995,tucker1979}. The vegetation indices includes Normalized gree (G), Excess green index (ExG), Excess blue index (ExB), Excess red index (ExR), Excess green-red index (ExGR), Normalized green-red difference index (NGRDI), Normalized green-blue difference index (NGBDI), Modified green-red vegetation index (MGRVI), Red-blue vegetation index (RBVI), Visible atmospherically resistant index (VARI), Green leaf index (GLI), and Red-green-blue vegetation index (RGBVI). Their means, medians, and 10th (p10) and 90th(p90) quantiles were computed. The formula of these vegetation indices is in Supplementary Table S1.

UAV flights were targeted weekly to capture images of the canopy throughout the growing season. Figure~\ref{fig:central_figure}e uses time-series data points to illustrate the number of UAV flights for each site-year. The generalized additive model (GAM) was used to fit a curve using at least four time points \cite{hastie2017generalized}. Figure \ref{fig:central_figure}e illustrates the number and the time of flight date in each site-year. In addition to the mean canopy area, peak of canopy area, area under the curve (AUC), maximum canopy growth rate, and days after planting (DAP) at the maximum growth rate were extracted from the fitted growth curve (Figure \ref{fig:central_figure}f).

\subsection{Statistical analysis}
The analysis was conducted using R in RStudio. The Pearson correlation function in the \textit{tidyverse} was used to calculate correlations between yield and other agronomic traits, and between traits extracted from the UAV pipeline and manual measurements \cite{wickham2019welcome}. Principal component analysis (PCA) was conducted to characterize correlations among variables and identify the major sources of variation in the dataset. 

Analysis of variance was performed using mixed-effects models, with genotype treated as a fixed effect for each response variable. Model assumptions were evaluated using residual diagnostics. Residual normality was assessed using the Shapiro--Wilk test, normal Q--Q plots, and residual histograms \cite{wickham2019welcome}, whereas homoscedasticity was evaluated using plots of residuals against fitted values. Potential outliers were identified through these diagnostic plots and removed only when supported by clear evidence.

The model was fitted using \textit{lme4} \cite{bates2015fitting} in R. Best linear unbiased estimates (BLUEs) were estimated using the following mixed linear model:
\begin{equation}
Y_{ijk} = \mu + (GE)_{ij} + R_k(E_j) + E_j + G_i +  \varepsilon_{ijk}
\end{equation}
where $Y_{ijk}$ is the observed phenotype of the $i$th genotype in the $j$th site-year and the $k$th replication, $\mu$ is the overall mean, $G_i$ is the fixed effect of genotype, $E_j$ is the random effect of site-year, $(GE)_{ij}$ is the random genotype-by-environment interaction effect, $R_k(E_j)$ is the random effect of replication nested within site-year, and $\varepsilon_{ijk}$ is the residual error.

Least Squares Means (LSMeans) were calculated for each genotype using the emmeans package \cite{lenth2019package}. The Multcomp and MultcompView packages were used to construct the distance matrix and identify significant differences \cite{hothorn2016package,graves2015package}.

 To determine the difference in seed yield between the CDC and PI sets, a new statistical model was used. Because the CDC and PI trials share no common genotypes or checks, we could not directly compare CDC genotypes with PI genotypes. 

The average yield performance of the two genotype sets was compared using a linear mixed model. The following mixed linear model was fitted:

\begin{equation}
Y_{ijkl}
=
\mu
+ E_i
+ S_j
+ (ES)_{ij}
+ G_{k(i)}
+ (SG)_{jk(i)}
+ B_{l(ij)}
+ \varepsilon_{ijkl},
\label{eq:experiment_comparison}
\end{equation}

where \(Y_{ijkl}\) is the observed yield of the \(k\)th genotype from the \(i\)th experiment in the \(j\)th site-year and the \(l\)th block; 
\(\mu\) is the overall mean; \(E_i\) is the fixed effect of experiment; 
\(S_j\) is the fixed effect of site-year; \((ES)_{ij}\) is the fixed 
experiment-by-site-year interaction effect; \(G_{k(i)}\) is the random effect of genotype nested within experiment; \((SG)_{jk(i)}\) is the random site-year-by-genotype interaction effect, with genotype nested within experiment; \(B_{l(ij)}\) is the random effect of block nested within the experiment-by-site-year combination; and \(\varepsilon_{ijkl}\) is the residual error.

 In the CDC–PI comparison, inference was restricted to the three specific environments in which the CDC trial was grown, and site-year (three levels) was therefore treated as a fixed effect. Genotype was treated as a random effect nested within experiment so that the experiment effect represented a comparison between the mean performances of the two genotype sets while accounting for variation among genotypes within each experiment, rather than relying only on plot-to-plot variation. This also reduced the influence of the different numbers of genotypes in the PI and CDC experiments.

Estimated lsmeans for the two experiments and the CDC--PI contrast were obtained using the \texttt{emmeans} package. The three site-years were weighted equally when calculating the estimated means, and denominator degrees of freedom were approximated using the Satterthwaite method.

Broad-sense heritability ($H^2$) was calculated using best linear unbiased prediction (BLUP), with genotype treated as a random effect. The variance components for genotype, environment, and their interaction were estimated using the \textbf{lme4} package \cite{bates2015fitting} according to the following model:

\begin{equation}
H^2 = \frac{V_G}{V_P}
= \frac{V_G}{V_G + \frac{V_{GE}}{e} + \frac{V_{\varepsilon}}{re}}
\end{equation}

where $H^2$ is broad-sense heritability, $V_G$ is the genotypic variance, $V_P$ is the phenotypic variance, $V_{GE}$ is the genotype-by-environment interaction variance, $V_{\varepsilon}$ is the residual variance, $e$ is the number of site-years, and $r$ is the number of replications.

To evaluate the reliability of the data consistency across site-years, the coefficient of variation ($CV$) was calculated as:

\begin{equation}
\mathrm{CV}(\%) = \frac{\sigma}{\mu} \times 100
\end{equation}

where $\sigma$ is the residual standard deviation from the model and $\mu$ is the mean of the observed trait.

\section{Result}
\subsection{Agronomic performances}

 Genotypes differed significantly in seed yield, seedling vigor, days to flowering, days to maturity, canopy height, and plant height. P-value table is in Supplementary Table S2. Figure~\ref{fig:combined_agronomic_traits} shows the mean agronomic and yield performance for all genotypes.  The average seed yield across six site-years was 973.89 kg/ha among 17 genotypes. The yield data showed that nine accessions had yield similar to the commercial check, Delta. PI505144 was the highest-yielding accession, had an average of 1429 kg/ha, and was significantly higher than three low-yielding accessions.

\begin{figure}[t]
    \centering
    \begin{subfigure}[b]{0.48\linewidth}
        \centering
        \includegraphics[width=\linewidth]{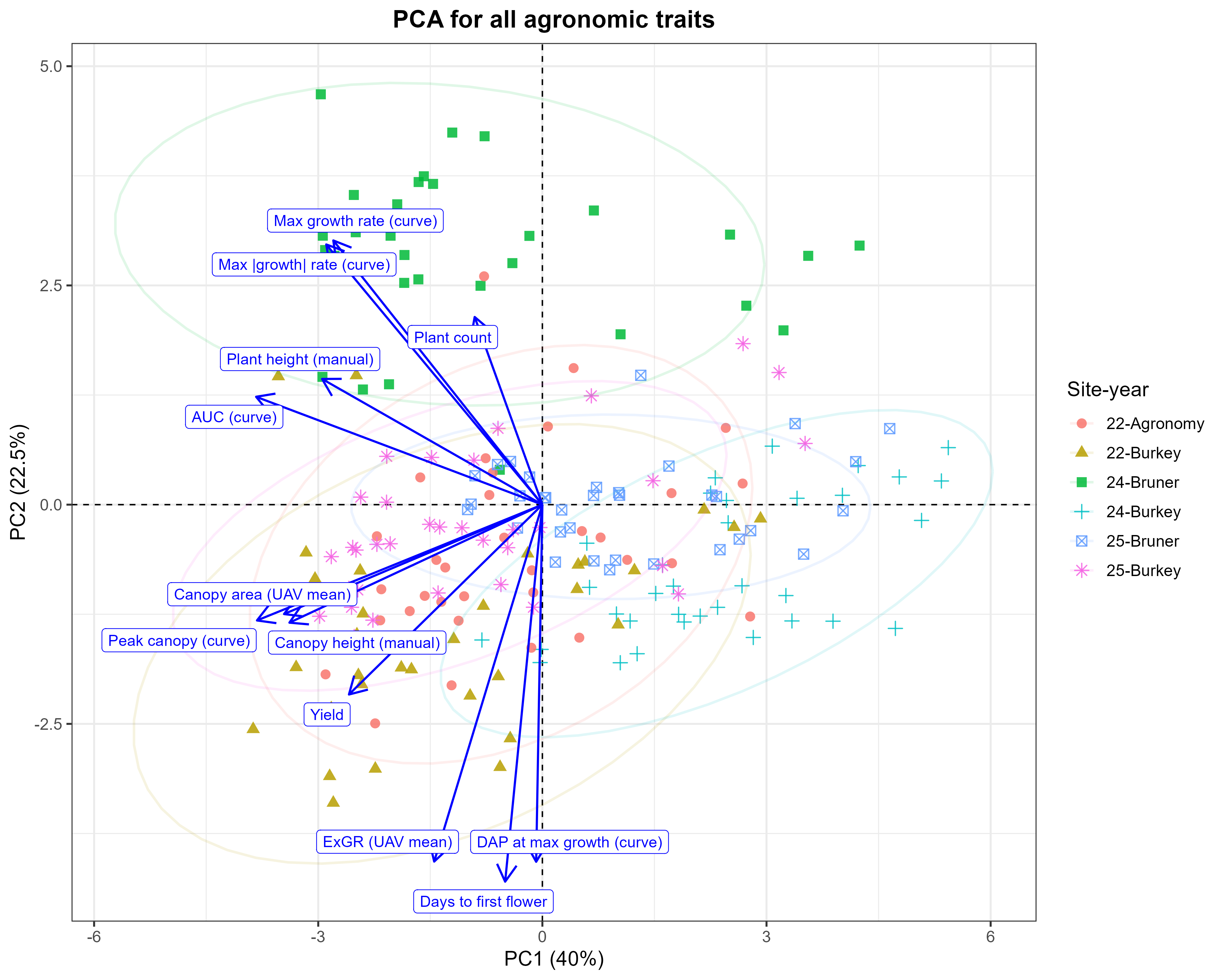}
        \caption{Phenotype PCA}
        \label{fig:pheno_pca}
    \end{subfigure}
    \hfill
    \begin{subfigure}[b]{0.48\linewidth}
        \centering
        \includegraphics[width=\linewidth]{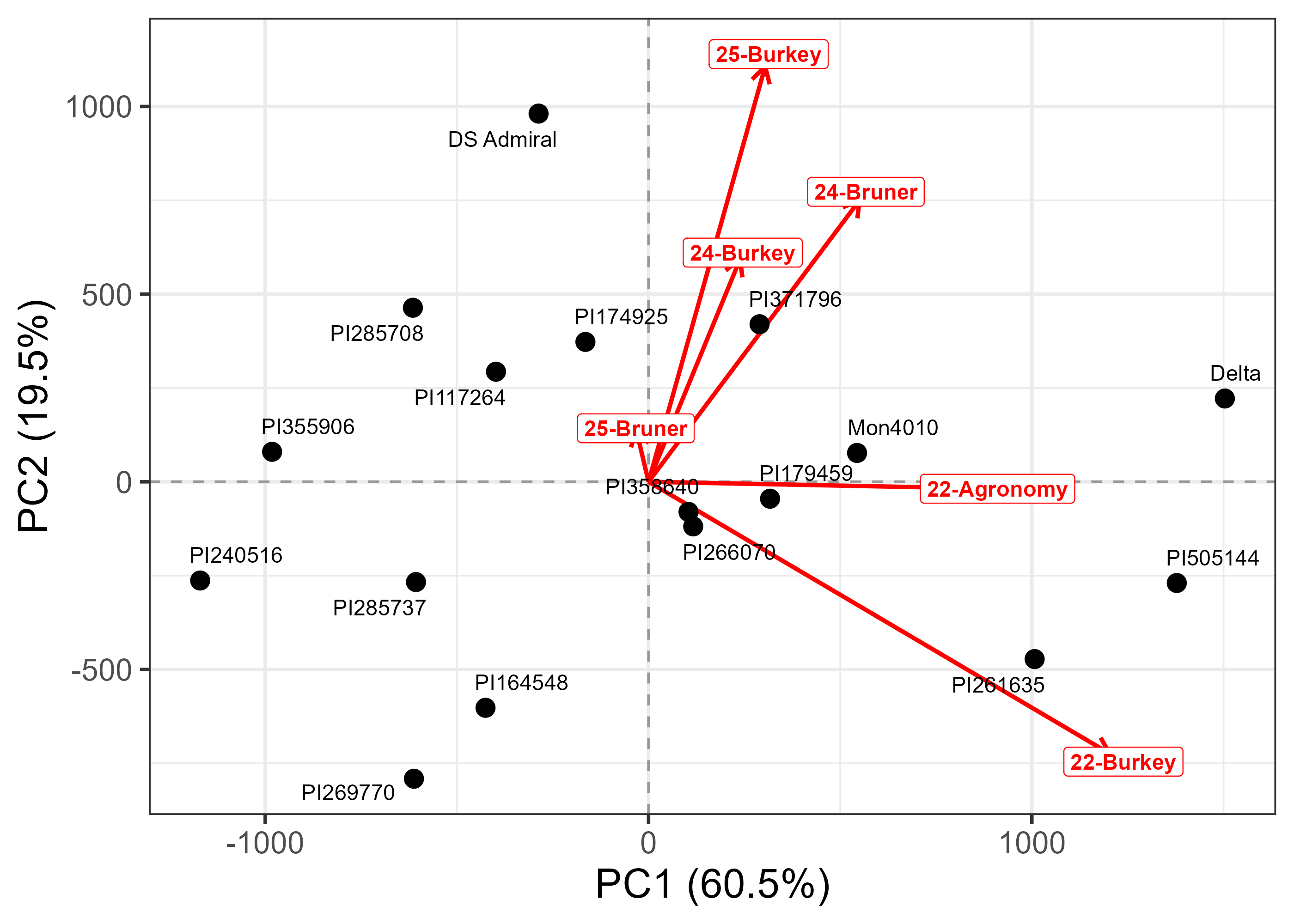}
        \caption{Genotype PCA}
        \label{fig:geno_pca}
    \end{subfigure}
    \caption{Principal component analysis (PCA) for field pea traits: a) Phenotype PCA for time-series canopy area, agronomic, and yield traits; and b) Genotype PCA mapping genotype performance across site-year distributions.}
    \label{fig:pca_combined}
\end{figure}

The Figure \ref{fig:pca_combined}b shows that genotypes varied across the site-years in yield. It shows the genotype consistency across site-years. Genotypes close to the central point indicate they performed more consistently across site-years, such as PI358640, PI266079, and PI179459. The high-yielding line PI505144 and Delta were more strongly associated with specific site-years. The yield results of mean seed yield ($\text{kg/ha}$) by genotype across each site-year shows in the Supplementary Figure S2. 

The results for DTF and DTM revealed a similar developmental pattern, segregating the 17 genotypes into distinct early-, mid-, and late-maturing groups (Figure \ref{fig:combined_agronomic_traits}c, d). The early-maturing group was characterized by a rapid vegetative phase, flowering in fewer than 45 days after planting (DAP) and reaching maturity within 75 days. The intermediate group flowered between 45 and 50 DAP (Figure \ref{fig:combined_agronomic_traits}c) and matured within 75 to 85 days (Figure \ref{fig:combined_agronomic_traits}d), whereas the late-maturing genotypes required more than 50 DAP to flower and exceeded 85 days to reach full maturity. In addition, most genotypes assigned to a specific DTF group fell into the corresponding DTM group; for instance, PI240516 and PI285737 were identified as significantly earlier in both flowering and maturity. The correlation between DTF and DTM is \textit{r} = 0.75(P $<$0.001). This indicates that the classification within a specific flowering window strongly predicted the final maturity timing.

Canopy and plant height data showed variation among the 17 genotypes (Figure \ref{fig:combined_agronomic_traits}e, f). Shorter-plant-height genotypes maintained low canopy heights. However, some high-plant-height genotypes exhibited low canopy heights; for example, PI117264 and PI164548 were long-stemmed yet had low canopy heights. This discrepancy indicates susceptibility to lodging among these specific tall lines. 

\subsection{Agronomic performance in Canadian elite varieties}
\begin{figure}
    \centering
    \includegraphics[width=0.75\linewidth]{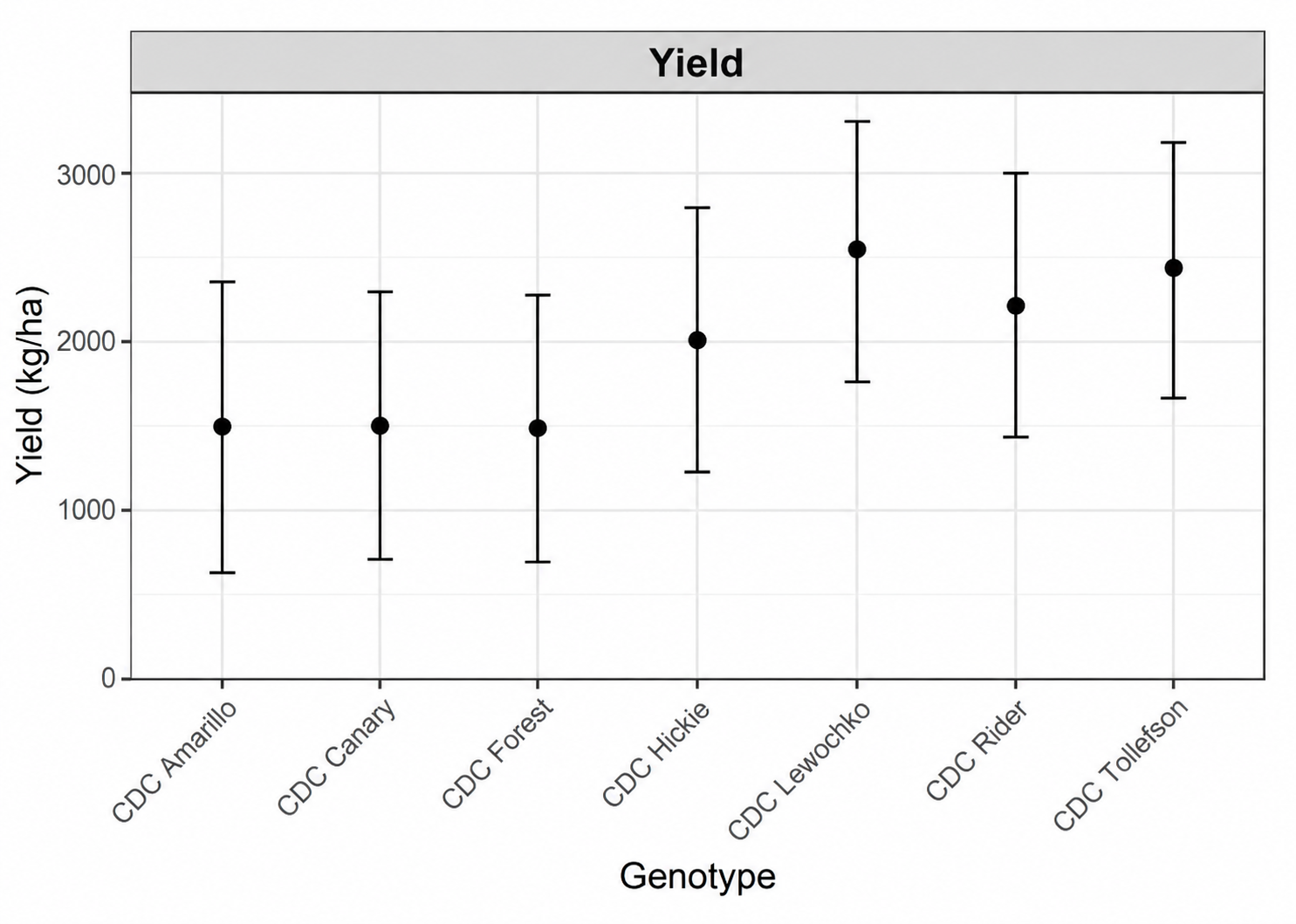}
    \caption{Seed yield (kg/ha) for Canadian elite varieties. Mean of two site-years in 2024 and 2025}
    \label{fig:CAD_yield}
\end{figure}

 For the Canadian elite varieties, seed yield showed a significant genotype-by-environment interaction between 24-Burkey (mean yield 628.80 kg/ha) and the other two environments (24-Bruner: 1897.60 kg/ha; 25-Bruner: 2052.28 kg/ha). This was due to severe early-season waterlogging in 24-Burkey, which stunted pea growth. To fairly evaluate the yield potential of the Canadian varieties in the Iowa environment, only the two representative site-years, 24-Bruner and 25-Bruner, were used, because 24-Burkey was strongly influenced by environmental conditions. Mean yield and agronomic performance for these varieties are shown in Figure\ref{fig:CAD_yield}. The average yield of the seven genotypes was 1961 kg/ha, and the highest-yielding genotype, CDC Lewochko, produced 2566 kg/ha.

 The results of PI set and CDC sets showed that the interaction between experiment and environment was significant (P$<$0.001); the yield advantage was not consistent across environments. The CDC genotype set yielded significantly more than the PI genotype set in 24 and 25-Bruner (P $<$ 0.0001) and showed no difference in 25-Burkey (Supplementary Note~1).

\subsection{UAV-derived traits and agronomic traits}

\begin{figure}[t]
    \centering
    \includegraphics[
        width=0.9\textwidth,
        height=0.85\textheight,
        keepaspectratio
    ]{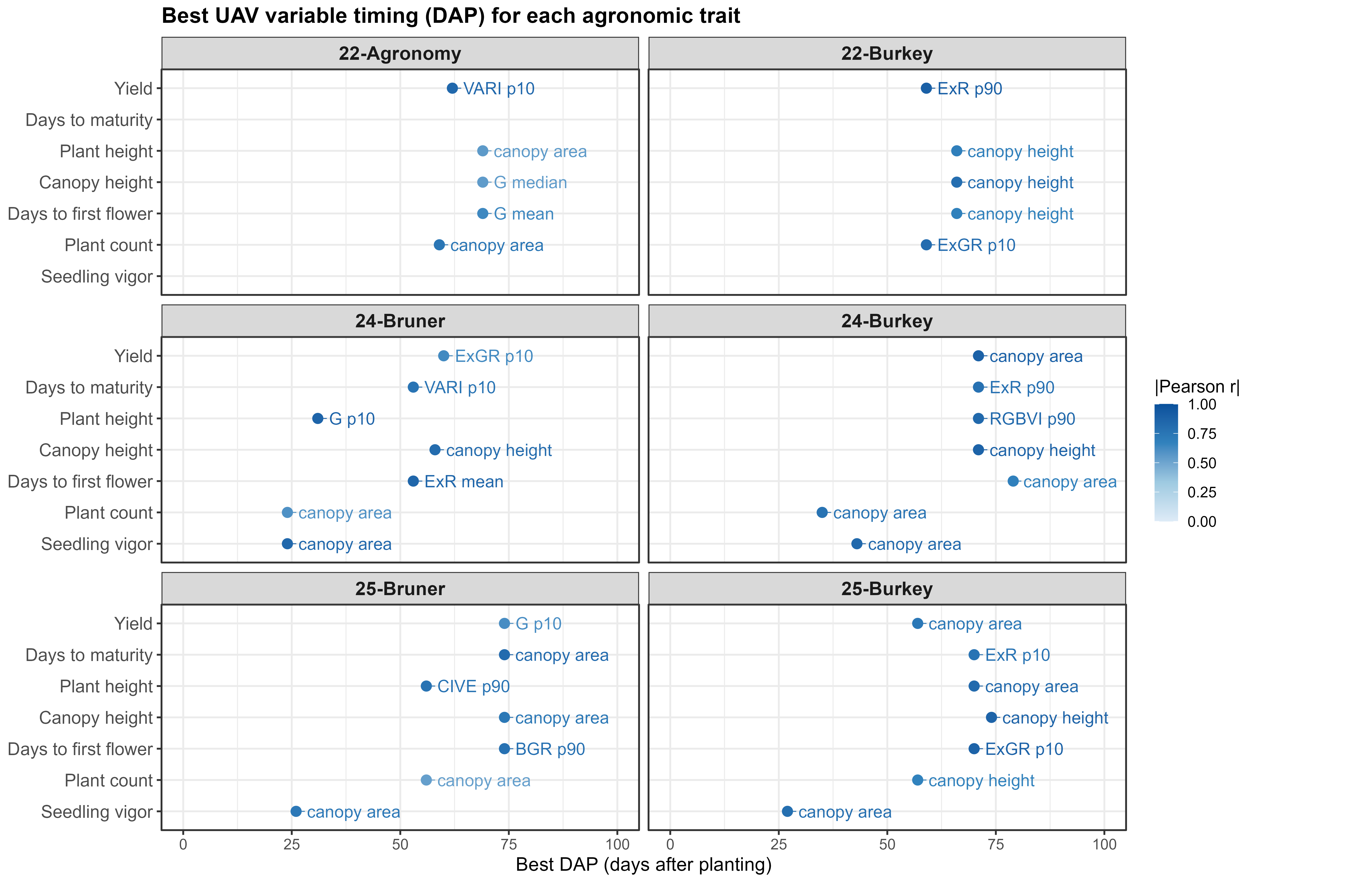}
    \caption{UAV-derived trait relevant to the agronomic traits. The x-axis represents the days after planting, and the y-axis lists agronomic traits. The blue dots represent the correlation of the UAV trait to the agronomic trait. Dot is darker, representing a higher correlation.}
    \label{fig:bestDAP_bestUAV}
\end{figure}

Correlation coefficients between all UAV traits at each time point and their corresponding ground measurements were calculated. Figure \ref{fig:bestDAP_bestUAV} illustrates the specific UAV traits and time points that exhibited the strongest correlations with manual agronomic traits. Among these, canopy area was the trait most highly correlated with seedling vigor across four environments, indicating that early-season UAV-derived canopy area effectively captures seedling vigor. Seedling vigor was visually evaluated by measuring seedling biomass volume three to four weeks after planting, which coincided with the date of the UAV image collection.
For other agronomic traits, no single UAV-derived trait demonstrated a consistent correlation across all environments; instead, the most highly correlated trait varied by site-year. Taking 25-Bruner as an example, canopy area at 25 days after planting (DAP) achieved a correlation of 0.65 with the seedling vigor trait, representing the strongest relationship observed. For canopy height, the UAV-derived canopy area at 75 DAP correlated most strongly with the ground-measured canopy height. Additionally, the greenness 10th quantile ($G_{p10}$) was identified as the most highly correlated UAV trait to grain yield ($r = 0.5$).Furthermore, canopy area was repeated multiple times with several agronomic traits across different growth stages and environments in the Figure \ref{fig:bestDAP_bestUAV}. It indicates that canopy area is the most relevant feature for assessing field pea agronomic performance and yield potential.

To validate that the UAV-derived canopy area is the most yield-relevant trait, the correlation coefficients between seed yield with all agronomic and UAV-derived metrics were calculated (Supplementary Figure S4). The result shows that the mean canopy area was the strongest correlated trait ($r$=0.63***) with final seed yield among all evaluated parameters. Consequently, canopy area was selected as the primary feature for subsequent time-series growth analysis.

Figure \ref{fig:central_figure}g shows that the canopy area increased progressively during the growing season, with increasing genetic variation from early to later DAPs. It demonstrates that the time-series data successfully illustrate canopy growth throughout the season, while the difference in canopy area among genotypes was greater over time.  

Principal component analysis for phenotype showed the trait associations across the whole dataset (Figure \ref{fig:pca_combined}a). Mean canopy area, peak canopy area, and manually measured canopy height were closely aligned with yield, indicating strong positive associations and more useful time-series variables relevant to the yield. These PCA results were also consistent with our correlation analysis. 

\begin{figure}
    \centering
    \includegraphics[width=0.8\linewidth]{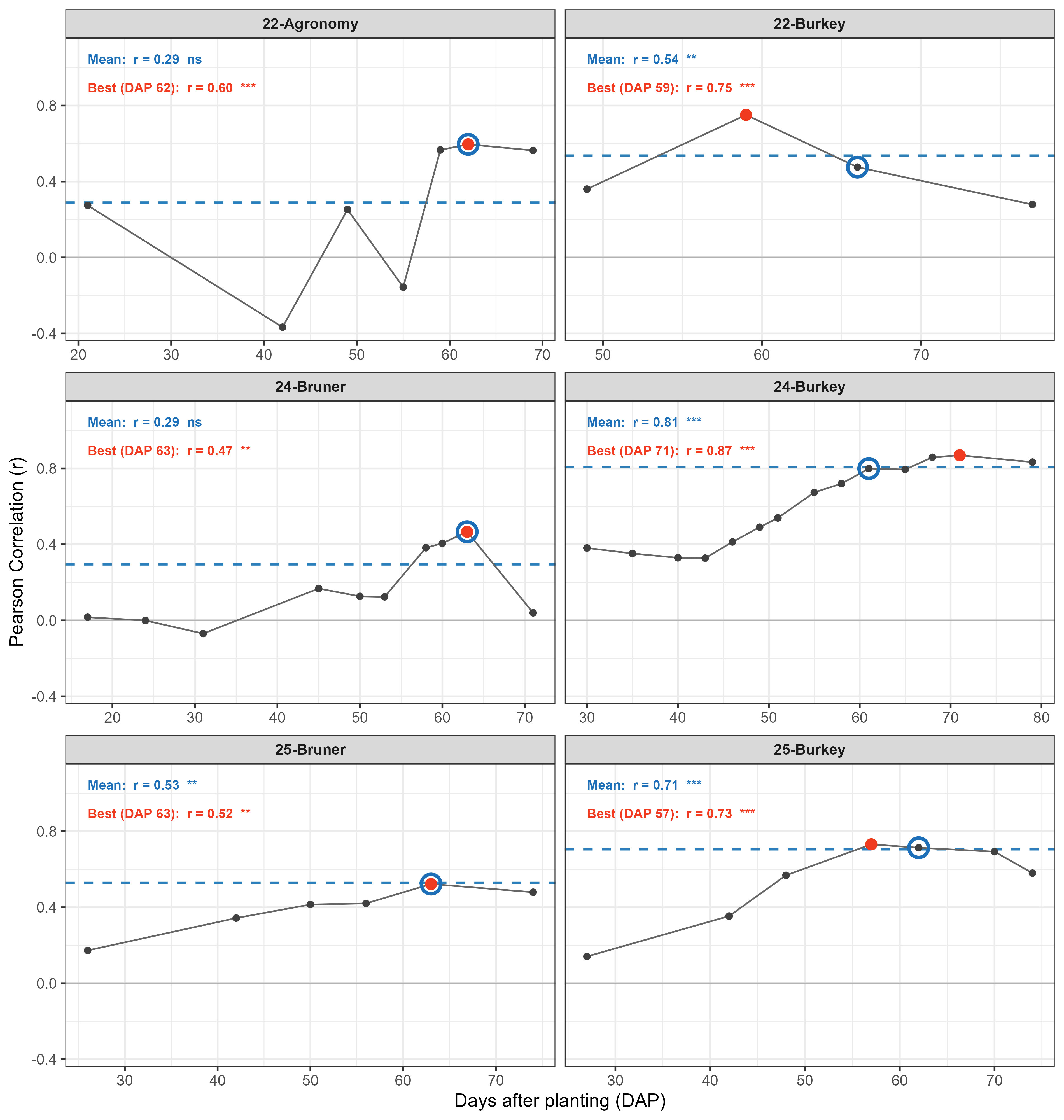}
    \caption{Comparison of the correlation between canopy area and yield throughout the growing season in each site-year. Each dot represents a correlation value at a specific DAP, and the solid line shows the variation in correlation within each site-year. The blue dashed line represents the averaged correlation across the season. The red dot indicates the flight date with the highest correlation. The blue circled dot indicates a flight during 60–65 DAP, at which the correlation was relatively close to, or at, the best single-date flight timing.}
    \label{fig:single_ts_area}
\end{figure}

The optimal timing of UAV flight for seed yield prediction was estimated by comparing the time-series mean with the time point most strongly correlated with it. As shown in Figure \ref{fig:single_ts_area}, the best single time point has a higher correlation than the time-series mean in five of the six site-years. The optimal DAP for a single flight fluctuated between 59 and 71 DAP depending on the site-year. A fixed target of 62 DAP (Figure \ref{fig:single_ts_area}, blue circle) was evaluated as a standardized alternative. Measurements at 62 DAP performed comparably to the time-series mean but did not consistently outperform it. Consequently, if only one operational flight is feasible for yield prediction, an acquisition window between 60 and 65 DAP is recommended.

Furthermore, evaluating the reliability of UAV-derived traits is essential for characterizing field pea performance across environments. The comparative analysis of coefficient of variation (CV) and broad-sense heritability ($H^2$) demonstrates that UAV-derived canopy traits offer more reliable and repeatable measurements than yield. Specifically, peak canopy ($H^2 = 0.91$, $\text{CV} = 14.83\%$) and mean canopy area ($H^2 = 0.87$, $\text{CV} = 13.46\%$) exhibited substantially higher heritability and lower phenotypic variability than grain yield ($H^2 = 0.72$, $\text{CV} = 25.69\%$; Supplementary Table S2). These results indicate that mean canopy area and peak canopy, as some of the most yield-relevant traits, can provide highly heritable and accurate selection criteria for field pea breeding programs.

\section{Discussion}

By evaluating field pea in multi-site, multi-year trials across central Iowa, this study demonstrated the crop's adaptation potential in the Midwest. The identified accessions exhibit genetic variation in seedling vigor, DTF, DTM, plant height, and seed yield, providing a genetic resource for future variety development in Iowa.
  The average yield of the highest genotype, PI 505144, was 1429 kg/ha, and the average across 17 genotypes was 980 kg/ha. They cannot compete with recent commercial varieties in agronomic traits. The challenges of the accessions in this experiment are that they are originally designed to represent the diversity of the species and its wild relatives as the first pea core collection \cite{coyne2005usda}. It was developed based on geographical origin and flower color, rather than for seed yield and agronomic purposes. In addition, these accessions lack modern pea variety traits, including the semi-leafless leaf type and powdery mildew resistance \cite{GRINGlobalProjectND}. Domesticated normal leafed type tends to have more severe lodging than the semi-leafless type \cite{stelling1989problems}; in addition, lodging could cause up to 74\% yield loss \cite{amelin2003scientific}. Powdery mildew is an air-borne fungal disease caused by the biotrophic pathogen \textit{Erysiphe pisi DC.} Infection affects the foliar, stem, pod, and seed of the plant, with yield losses that may reach 50\% \cite{warkentin1996fungicidal}. During field observations, we found powdery mildew in all accessions except the DS Admiral check.
In comparison, Canadian elite varieties exhibit superior standability and disease resistance (Section 2.1). In the results, the CDC trial out-yielded the PI trial in two of the three environments (24- and 25-Bruner), with no significant difference in 25-Burkey. Because the two trials shared no common genotypes or checks, this represents a trial-level comparison rather than a direct genotype-level comparison (Supplementary Note~1). The CDC Lewochko was the highest-yielding line in this study, with 2566 kg/ha, but produced a lower yield than it did in the variety testing trial, with 3730 kg/ha \cite{Tollefson}. These results indicate a gap between pea production and pea varieties bred for the Midwest U.S. In the future, these Canadian lines can serve as elite checks for evaluating and selecting advanced lines within the breeding population. In addition, they can be used as parents crossed with the selected GRIN lines for pea breeding in the Midwest. 

Early DTF and DTM indicate that pea suits for the double-cropping system. The earliest pea genotype had an average maturity of 75 days. Following an early-maturing pea genotype with sorghum or millet allows for a double-cropping system. A recent three-year study (unpublished) from Iowa State University planted the millet from mid-June to late-June and harvested it in mid-October. With this information, peas can be planted in early to mid-April and have an 80-90-day growing season through July. Then, millet can be planted in the pea field and grown through the late summer months for a mid-October harvest. By improving land-use efficiency and farm profitability, a pea–cereal double-cropping system offers an environmentally and economically sustainable production model \cite{borchers2014multi}. However, there is often significant rainfall in Central Iowa from early April to May, which can delay planting or increase flood risk. Pea is susceptible to waterlogging \cite{ploschuk2018waterlogging}, which requires consulting field conditions based on the soil topography map.

Breeding is essential to adapt pea production to the Midwest; however, breeding programs must evaluate thousands of plots to make selections. Measuring in-field agronomic traits and manually selecting elite genotypes require substantial labor, time, and resources. To address this bottleneck, UAV-derived measurements provided a high-throughput approach that combined canopy color (vegetation indices) with canopy structure (area and height). In this study, we found that canopy area was the most useful UAV-derived trait, as it consistently correlated with seedling vigor and was the most relevant trait for pea yield. Canopy development has been considered a yield-component trait, related to canopy closure and weed suppression \cite{wall1996wild}. The higher-yielding genotype has a larger leaf area index and greater biomass accumulation during the flowering stage, both of which are strongly correlated with canopy area in the UAV image \cite{quiros2019phenotyping}. Previous studies also compared the correlation coefficients between yield and the vegetation index, and between canopy area in green and yellow peas, and the canopy area performed on par with the top-performing vegetation indices \cite{zhang2021crop}. Another study in winter wheat compared attention scores in the transformer model of yield prediction from vegetation indices versus canopy height and canopy area. The results also showed that canopy area and height contributed more important features than vegetation indices \cite{ge2025winter}. However, most PI genotypes in this study ended with lodging in the late season, and canopy closure occurred less uniformly. In the elite breeding population that all reach full canopy and lodging tolerance, canopy coverage features can saturate during peak growth, so UAV-derived structural traits such as canopy height and canopy volume, together with vegetation indices, may provide better discrimination \cite{shen2022optimizing}.

With time-series UAV phenotyping across all stages, this study classified a specific time point for traits of interest. Early canopy-area measurements were relevant to seedling vigor; in mid-season, vegetation index relevant to the days to flowering and days to maturity, UAV height relevant to canopy, and plant height. In the late season, multiple traits presented a strong correlation with yield. Our findings indicate that the best time window for predicting yield is 60-65 DAP (r=0.67). Previous studies predicted yield and the harvest index (the ratio of seed yield to above-ground biomass) and seed yield using vegetation indices, canopy cover, and canopy surface models from UAV images across varying stages \cite{liu2024ensemble,ji2024faba}; The results showed that the podding stages for pea were the best estimation (coefficient of determination ($R^2$) = 0.54) for the harvest index estimations than branching, budding, and flowering, and similar performance to the whole stage \cite{ji2024faba}. In addition, canopy area showed a non-significant correlation with yield in the early season but a strong, highly significant positive correlation (r = 0.85) in the mid-to-late growth stages \cite{zhang2021crop}.  Compared with single-time-point measurements, which focus on a single trait at a specific developmental stage, the time-series UAV approach in field pea captures the dynamic progression of crop development and tracks multiple plant traits across the season.  This interpretation is consistent with recent UAV phenotyping studies stated that no single trait consistently performs better across all growth stages, and that both vegetation indices and canopy area and height traits have varying effects as the crop grows \cite{ge2025winter,borra2020closing}. 

UAV phenotyping measurements can capture multiple agronomically relevant traits over time and have demonstrated precision and reliability. UAV-derived canopy traits such as peak and mean canopy area have high broad-sense heritability and low coefficients of variation for environmental effects. UAV-derived plant height estimates have shown strong agreement with ground measurements and high broad-sense heritability, and UAV platforms can provide heritability estimates that are at least comparable to, and sometimes better than, ground-based measurements across crop cycles in wheat breeding \cite{hassan2019accuracy}. These findings support the applicability of UAV phenotyping to pea breeding.

Field pea breeding in the Midwest is still in its early stages; the current breeding population retains substantial genetic diversity. The program draws on germplasm from the pea core collection, while only a limited number of advanced-generation lines have been developed and selected under Midwest environmental conditions. The absence of a large set of well-adapted advanced breeding lines constrains direct selection based on yield alone. Since yield is a complex quantitative trait strongly influenced by genotype-by-environment interaction, yield-based selection may be less reliable in this context. Consequently, incorporating indirect selection based on UAV-derived traits may improve selection efficiency and accelerate genetic gain.

\subsection{Conclusion}
In conclusion, this study evaluated field pea agronomic performances over three years in Iowa. This study found that PI accession trials exhibit genetic diversity in phenology and canopy traits, whereas the CDC elite varieties show more realistic potential yield. Short-maturity accessions planted in April can fit within the tight July planting window required for double-cropping with sorghum or millet in Midwest rotations. Local adaptation and variety development via long-term breeding are demanded to meet the requirements of pea production. This study demonstrated UAV-derived traits as a high-throughput breeding solution. UAV time-series phenotyping, specifically canopy area, as a method to overcome phenotyping bottlenecks in diverse germplasm to bring a reliable, yield-relevant trait that can be used in phenotypic selection to accelerate genetic gain in Midwest field pea breeding programs.

\section{Conflict of interest}
The author(s) declared no conflict of interest. 

\section{Acknowledgment}
We express our gratitude to Benjamin Marlow and Brian Scott for their valuable field management support and technical assistance throughout the trials. We are also thankful to the staff in Singh-Singh labs, as well as the graduate students and undergraduates involved in harvesting, imaging, and data collection. 

\section{Funding}
This research was supported by the Raymond F. Baker Center for Plant Breeding at Iowa State University.

\section{Figures}

\begin{figure*}[p]
    \centering

    \begin{subfigure}[t]{0.47\textwidth}
        \centering
        \includegraphics[width=\textwidth]{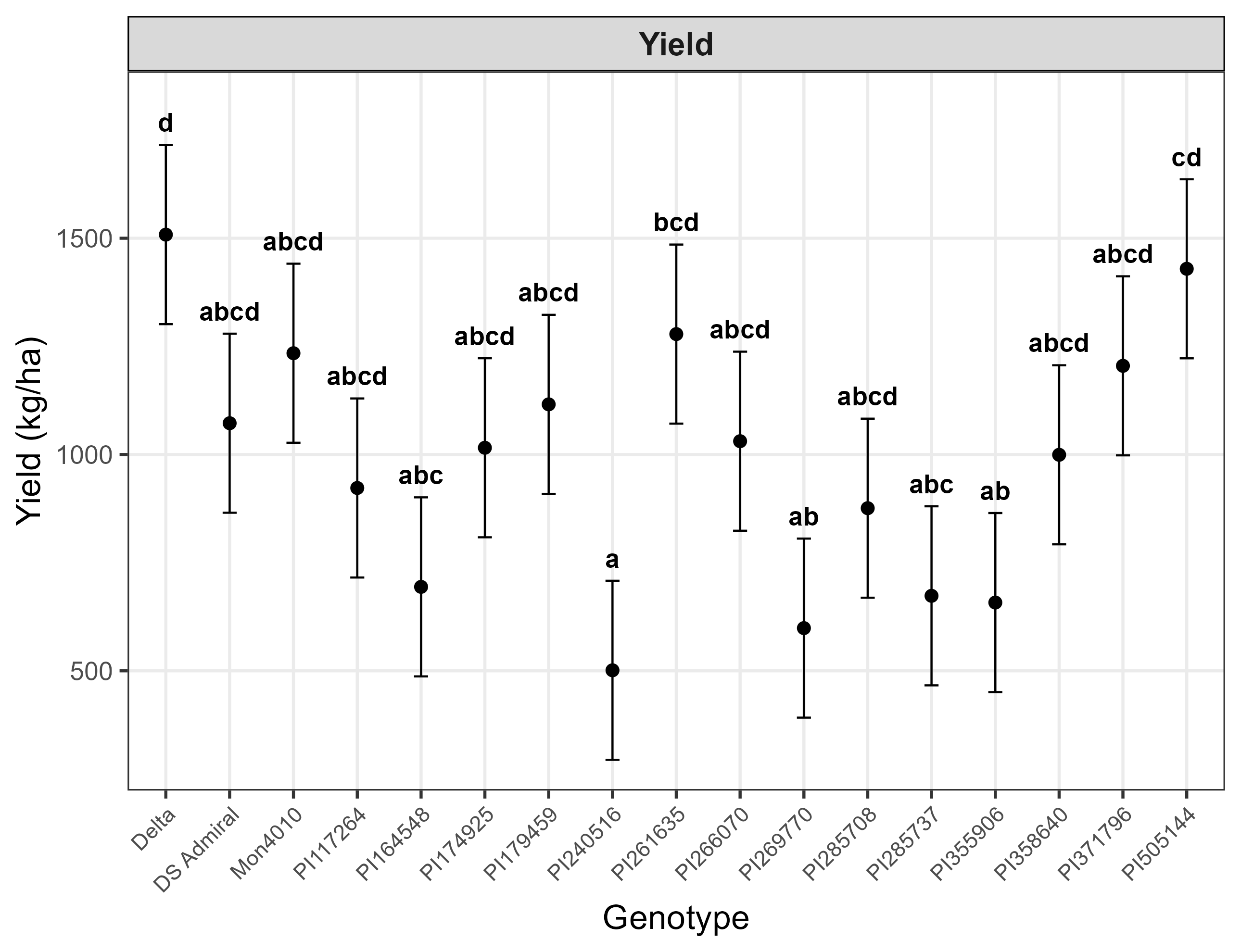}
        \caption{Grain Yield}
        \label{fig:sub_yield}
    \end{subfigure}
    \hfill
    \begin{subfigure}[t]{0.47\textwidth}
        \centering
        \includegraphics[width=\textwidth]{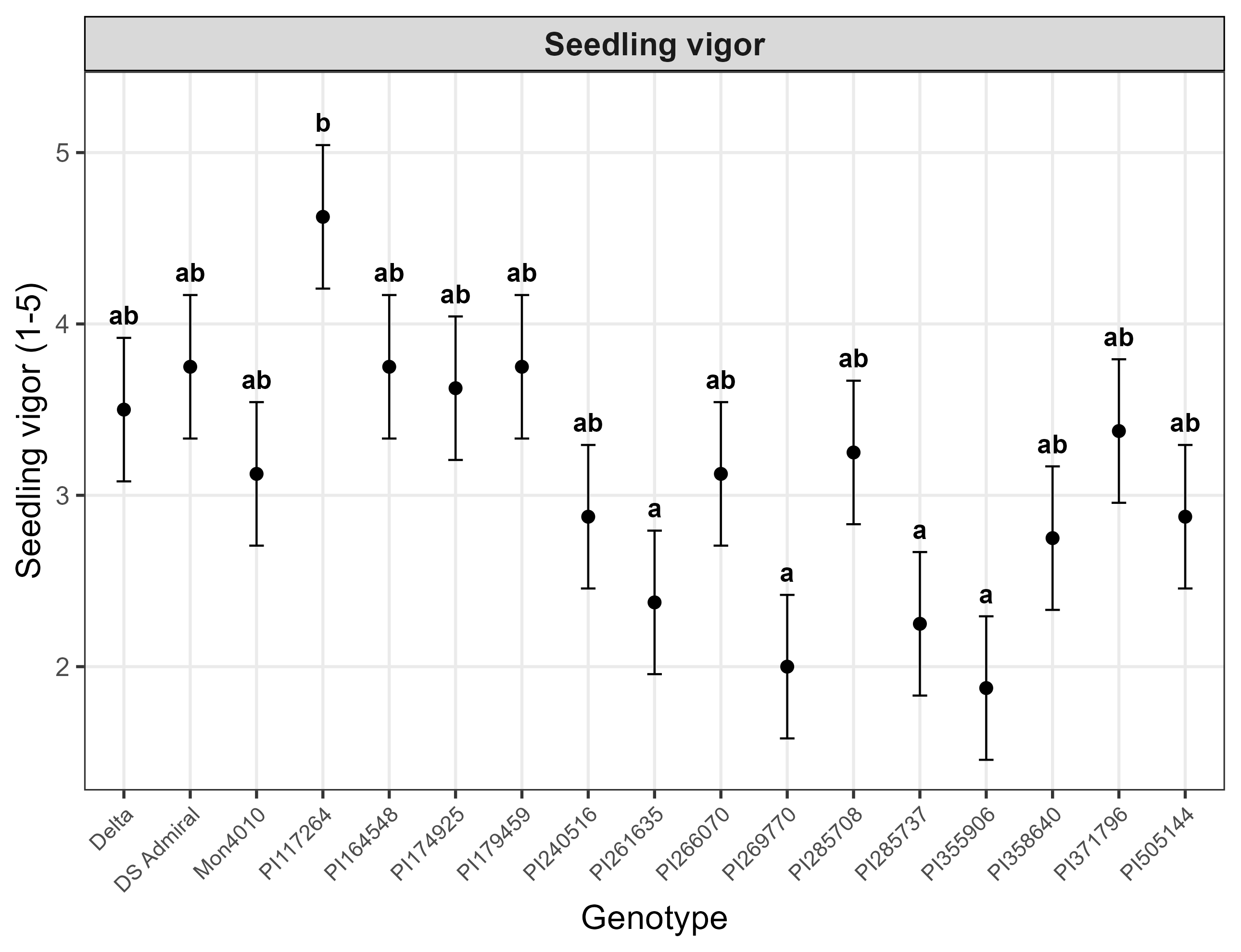}
        \caption{Early Seedling Vigor}
        \label{fig:sub_vigor}
    \end{subfigure}

    \vspace{0.15cm}

    \begin{subfigure}[t]{0.47\textwidth}
        \centering
        \includegraphics[width=\textwidth]{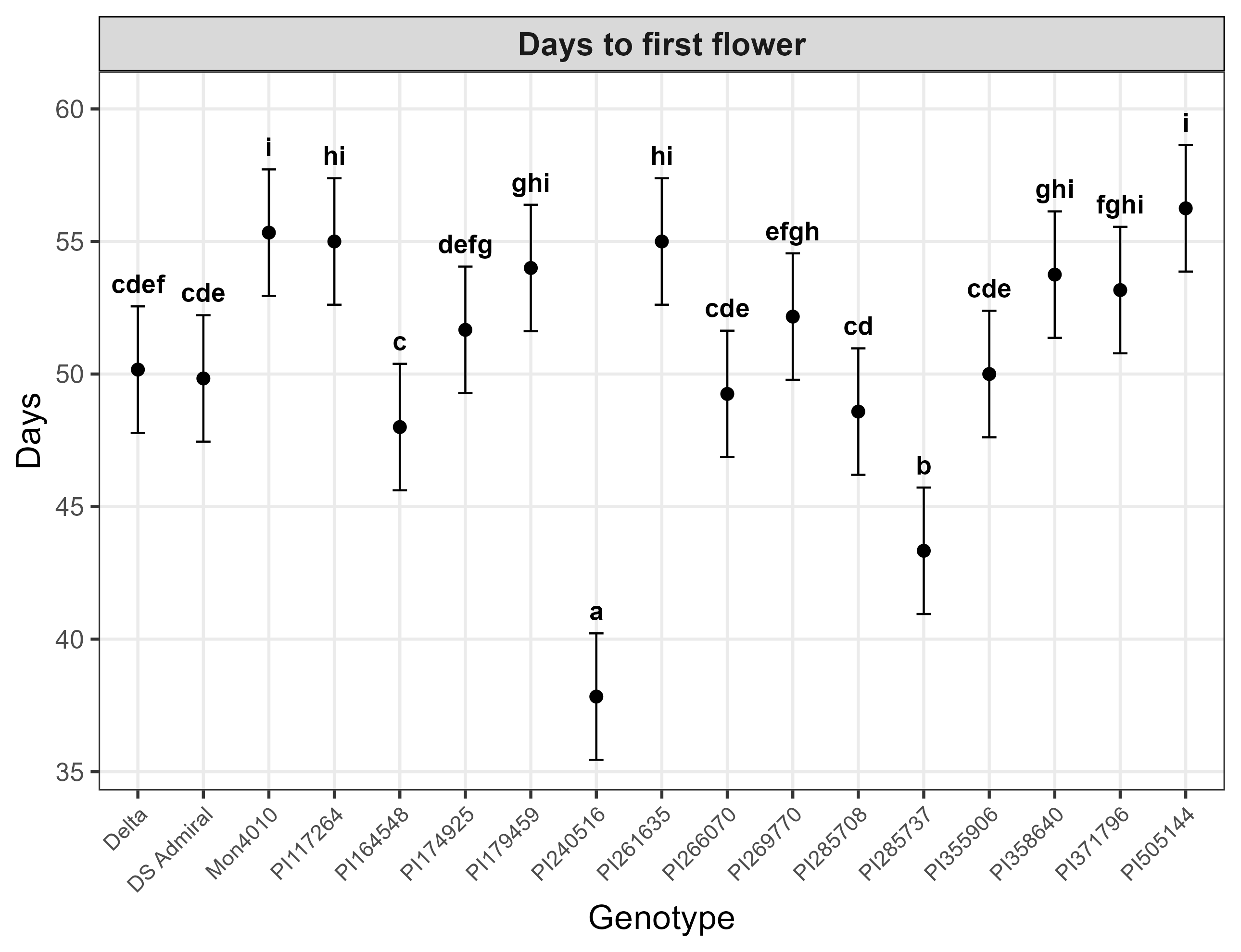}
        \caption{Days to Flowering (DTF)}
        \label{fig:sub_dtf}
    \end{subfigure}
    \hfill
    \begin{subfigure}[t]{0.47\textwidth}
        \centering
        \includegraphics[width=\textwidth]{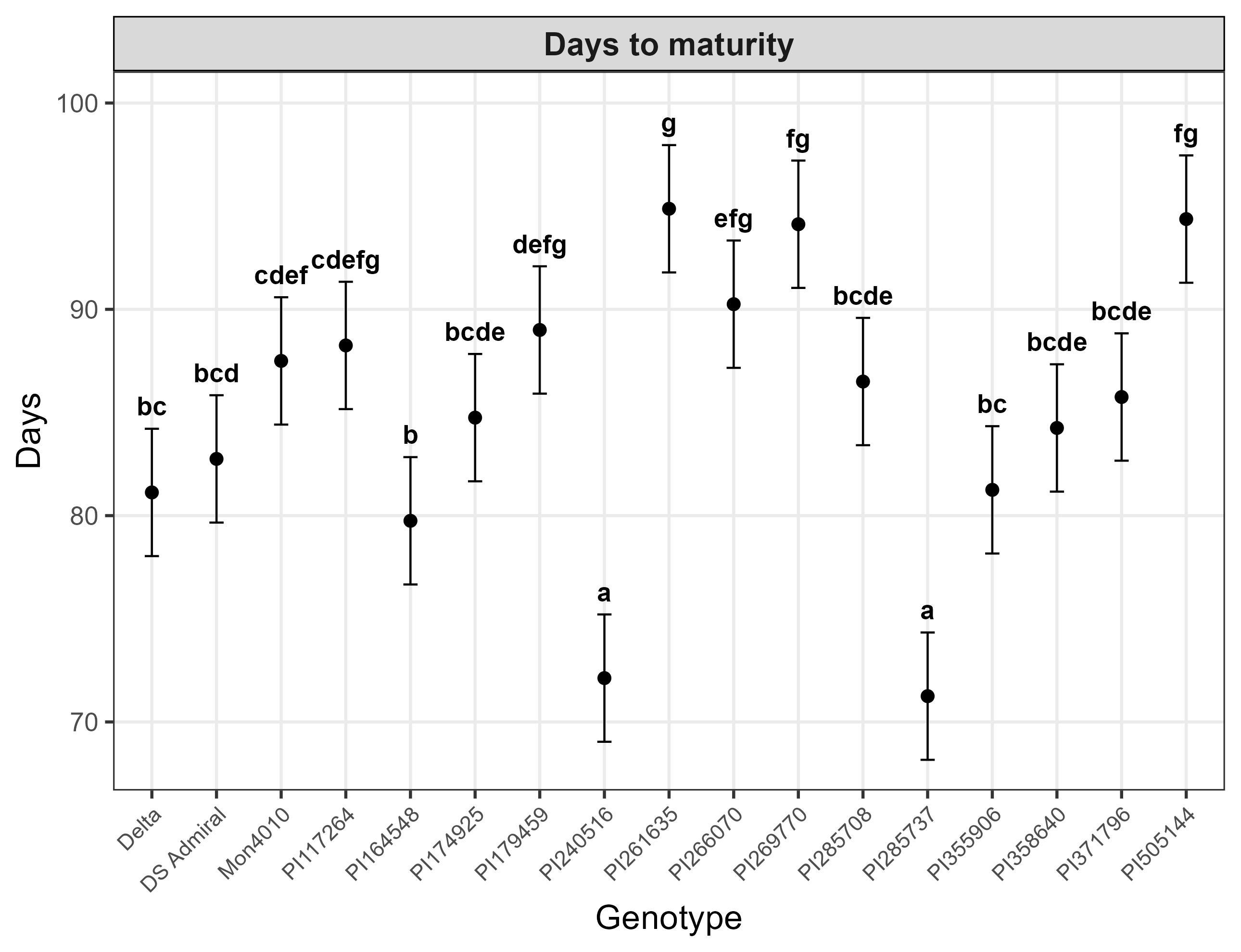}
        \caption{Days to Maturity (DTM)}
        \label{fig:sub_dtm}
    \end{subfigure}

    \vspace{0.15cm}

    \begin{subfigure}[t]{0.47\textwidth}
        \centering
        \includegraphics[width=\textwidth]{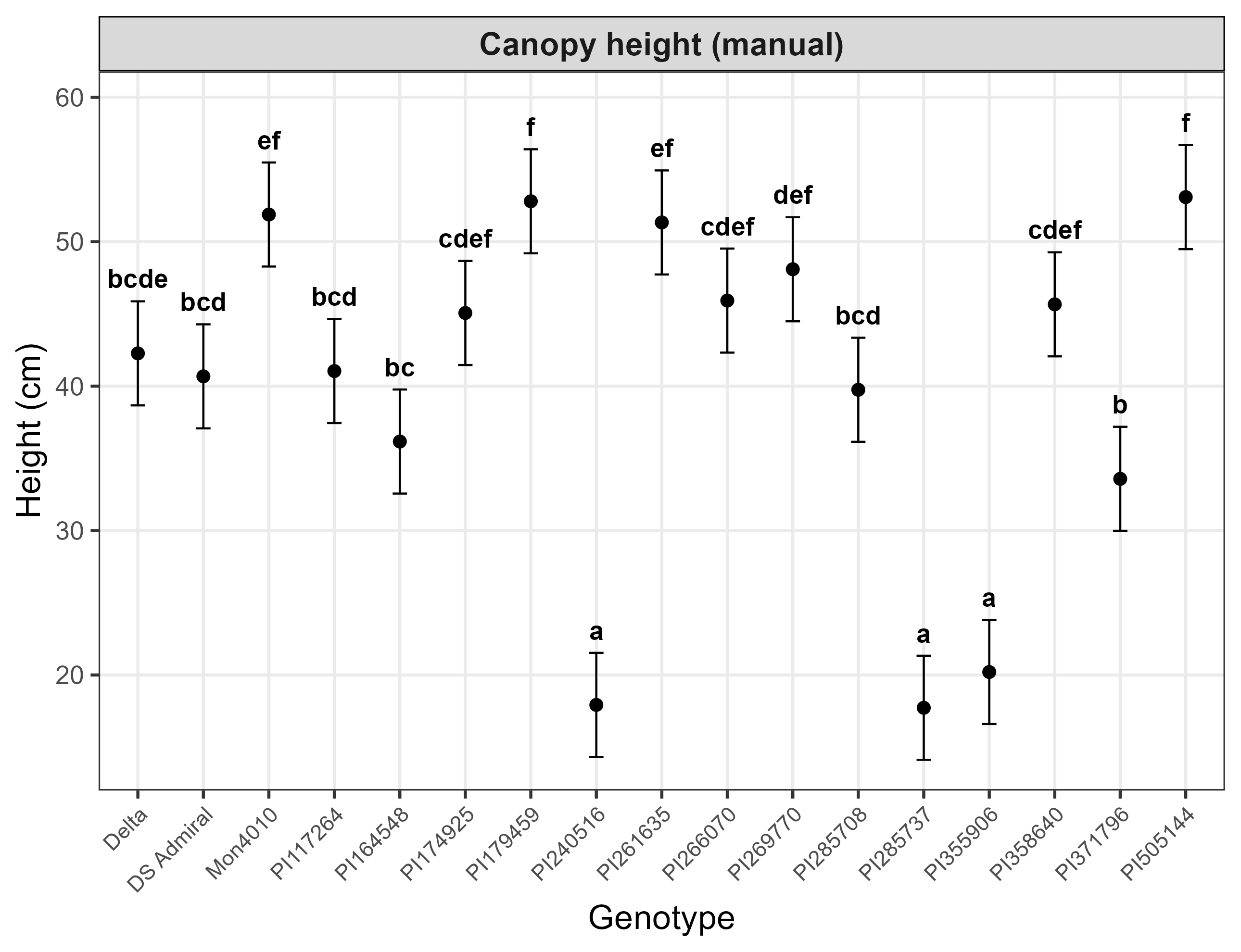}
        \caption{Canopy Height}
        \label{fig:sub_canopy_height}
    \end{subfigure}
    \hfill
    \begin{subfigure}[t]{0.47\textwidth}
        \centering
        \includegraphics[width=\textwidth]{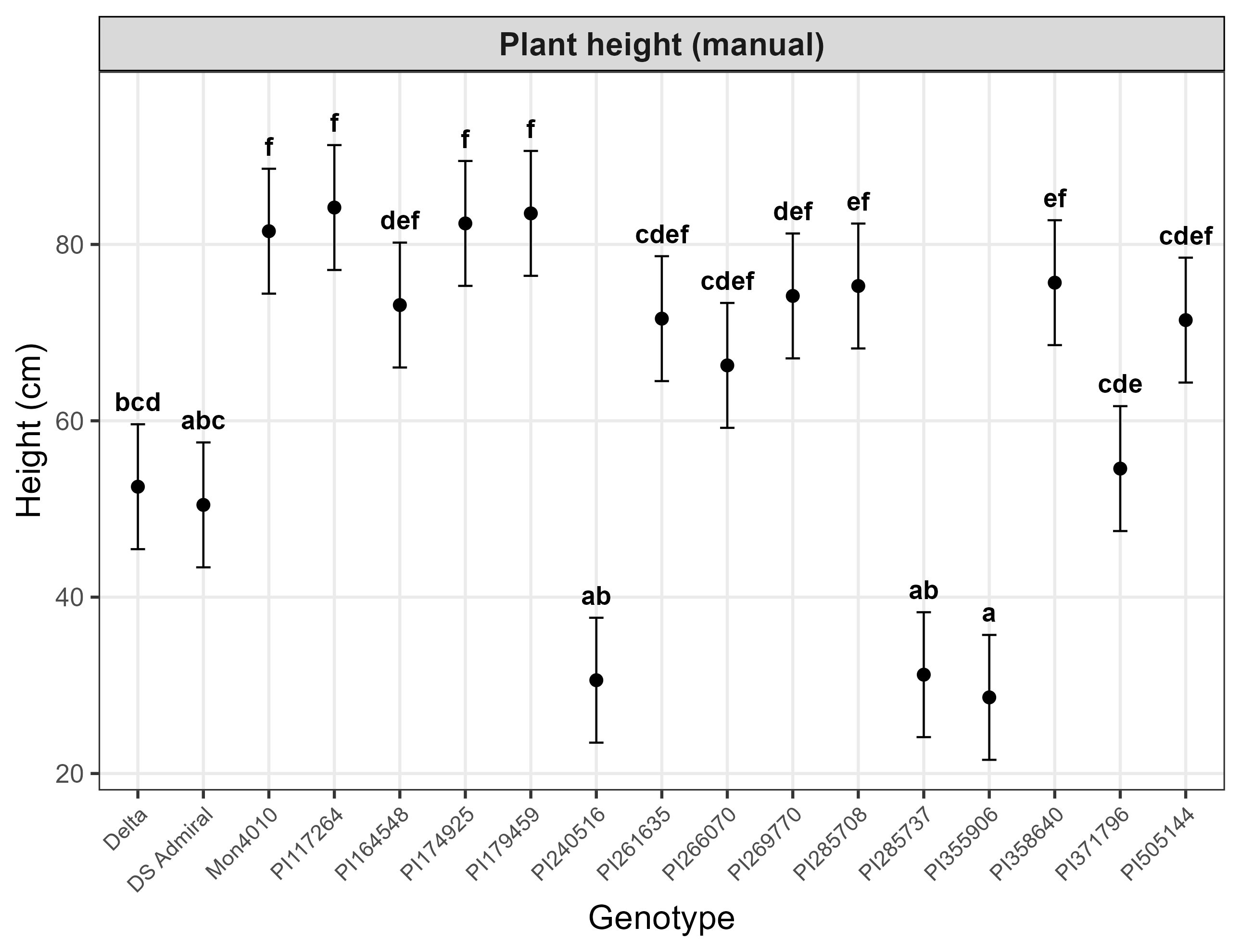}
        \caption{Plant Height}
        \label{fig:sub_plant_height}
    \end{subfigure}

    \caption{Comparison of manual agronomic traits and yield performance across pea genotypes, including (a) grain yield, (b) early seedling vigor, (c) days to first flower, (d) days to maturity, (e) canopy height, and (f) plant height.}
    \label{fig:combined_agronomic_traits}
\end{figure*}

\begin{figure*}[t]
    \centering

    \begin{subfigure}[t]{0.46\textwidth}
        \centering        
        \includegraphics[width=\linewidth]{figure/yield.png}
        \caption{Grain yield}
        \label{fig:PI_yield}
    \end{subfigure}
    \hfill
    \begin{subfigure}[t]{0.46\textwidth}
        \centering  
        \includegraphics[width=\linewidth]{figure/canopy_height.png}
        \caption{Canopy height}
        \label{fig:canopy_height}
    \end{subfigure}

    \vspace{0.3cm}

    \begin{subfigure}[t]{0.46\textwidth}
        \centering  
        \includegraphics[width=\linewidth]{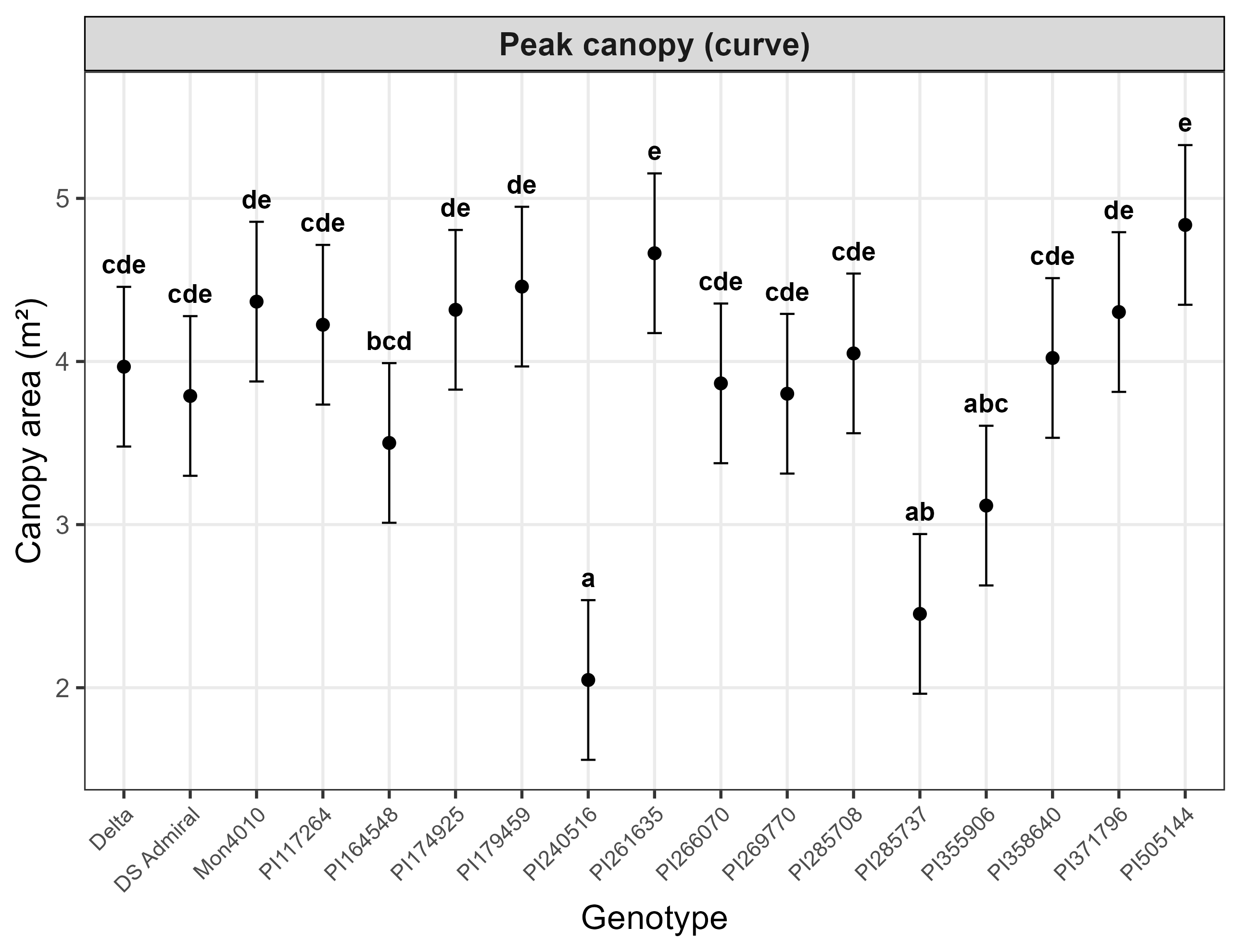}
        \caption{Peak canopy}
        \label{fig:peak_canopy}
    \end{subfigure}
    \hfill
    \begin{subfigure}[t]{0.46\textwidth}
        \centering        
        \includegraphics[width=\linewidth]{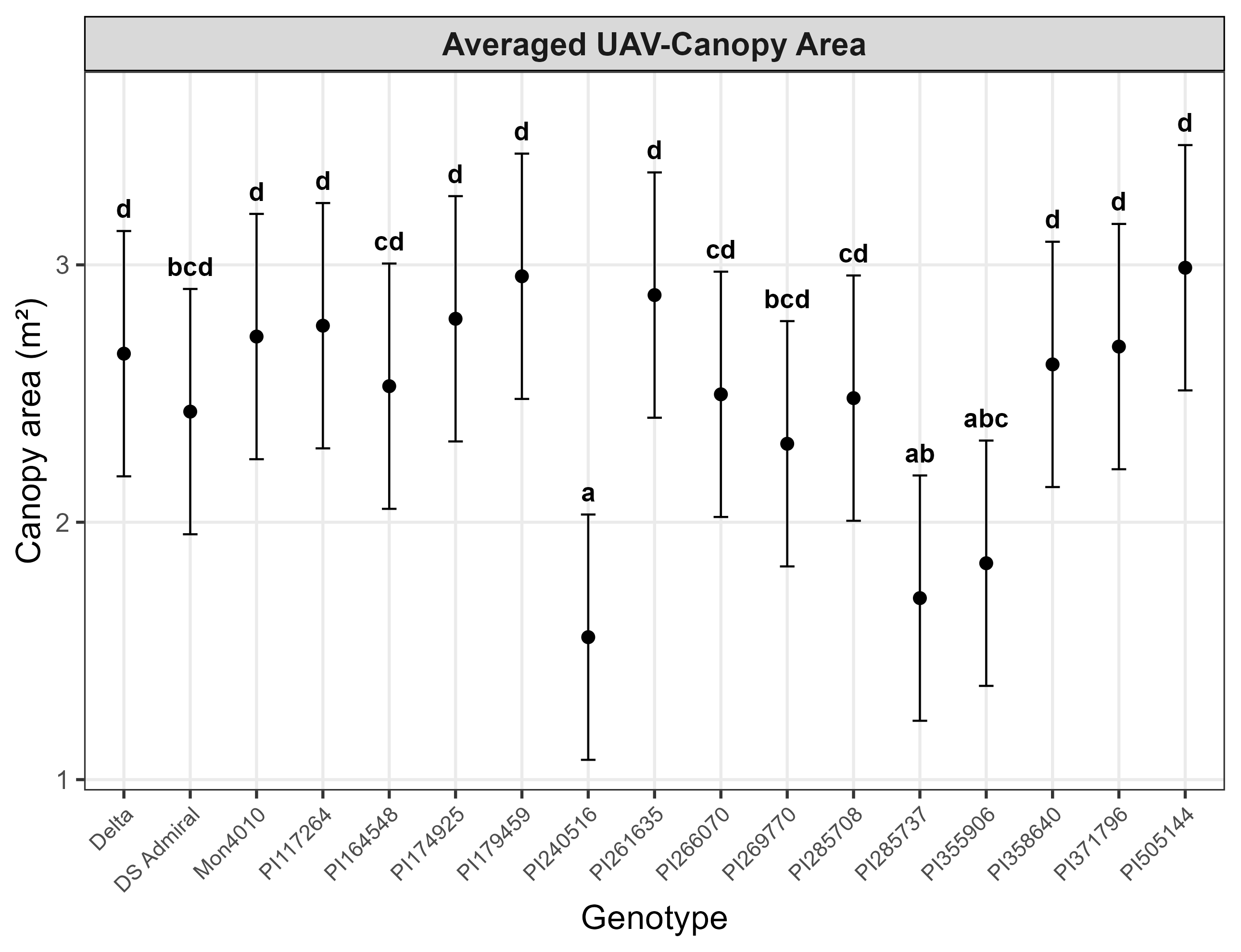}
        \caption{Averaged canopy area}
        \label{fig:average_canopy_area}
    \end{subfigure}

    \caption{Comparison of agronomic and UAV traits across site-years: (a) grain yield, (b) canopy height, (c) peak canopy, and (d) averaged canopy area.}
    \label{fig:agronomic_uav_traits}
\end{figure*}

\clearpage
\printbibliography

\end{document}


\begin{titlepage}
\centering

\vspace*{1.5cm}

{\Large \textbf{Supplementary Information}\par}

\end{titlepage}

\suppnote{Evaluating elite Canadian varieties performance in the Midwest}{supp_can_yt}

\section{Supplementary Figures}
\begin{figure}[htbp]
    \centering
    \includegraphics[width=\textwidth]{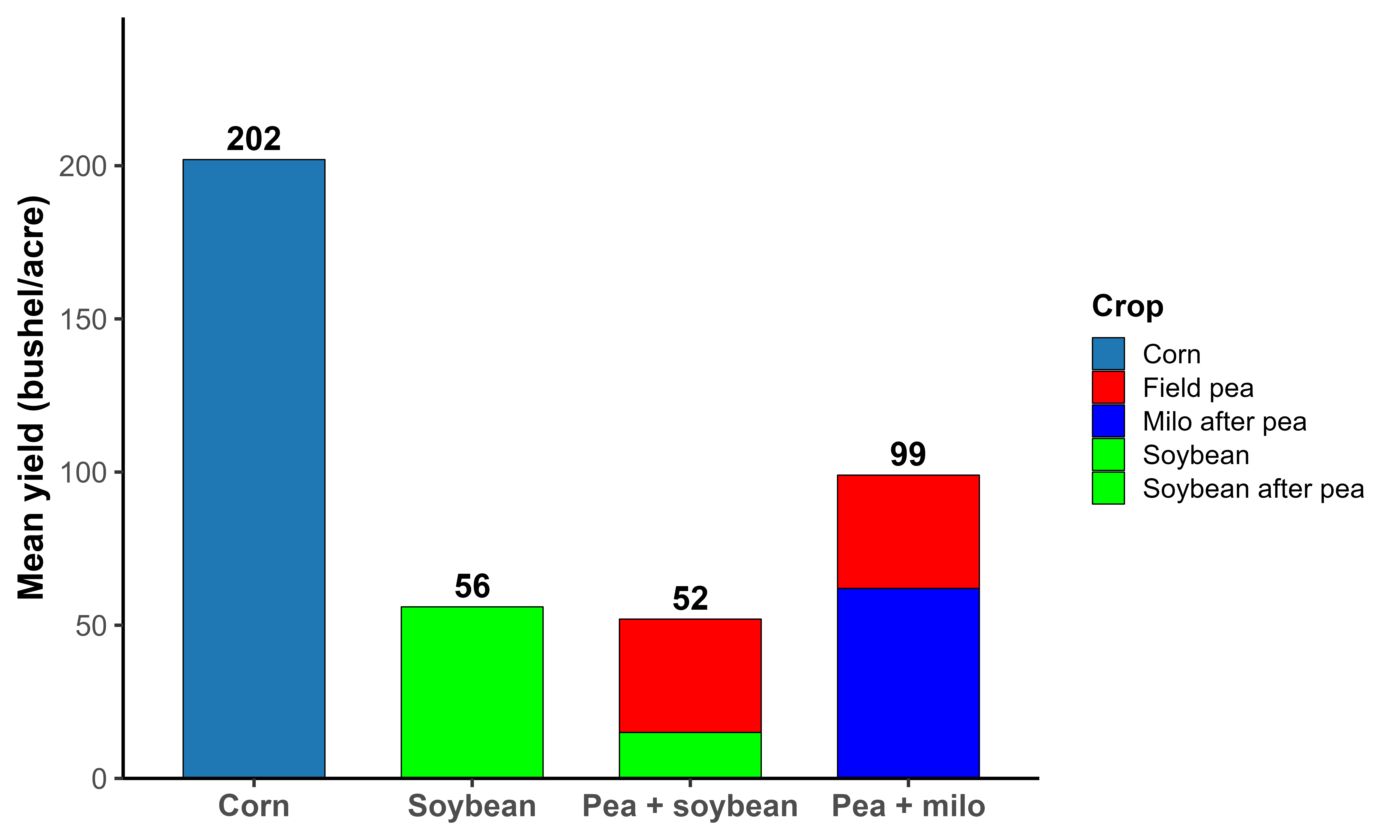}
    \caption{The averaged yield of different cropping system of pea plus milo (sorghum), pea plus soybean, soybean monoculture, and corn monoculture. The data source was pull from \cite{Fawcett2006DoubleCroppedFieldPea}. }
    \label{supp_fig:doublecrop}
\end{figure}

\begin{figure}
    \centering
    \includegraphics[width=1\linewidth]{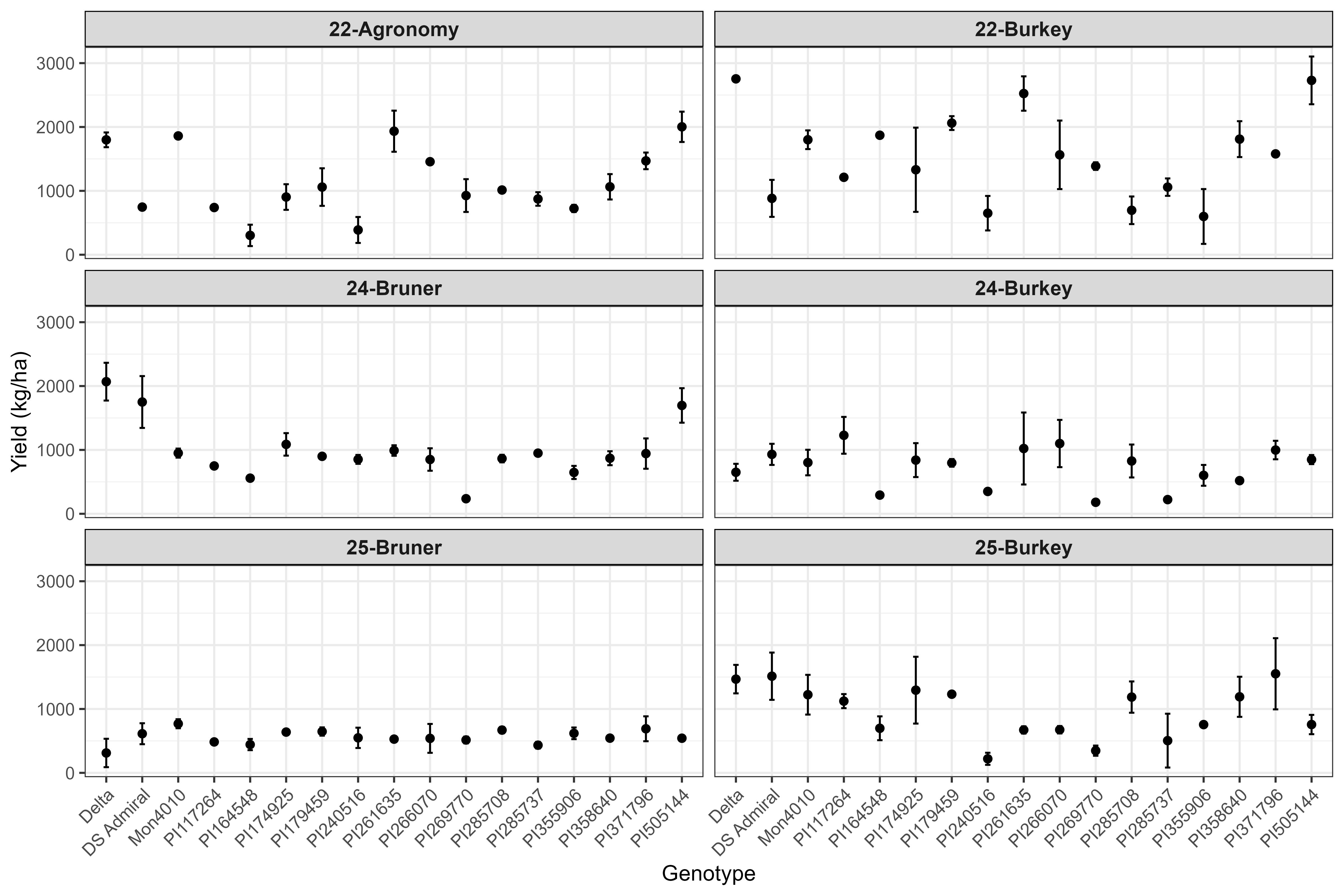}
    \caption{Mean seed yield ($\text\{kg/ha\}$) by genotype across each site-year. The trend of genotype seed yields varied across site-years. }
    \label{fig:yield_siteyear}
\end{figure}
    
    
    
    
    


\begin{figure}[htbp]
    \centering
    \begin{subfigure}[b]{0.32\textwidth}
        \centering
        \includegraphics[width=\textwidth]{figure/CAD_yield.png}
        \caption{Grain yield}
        \label{fig:cad_yield}
    \end{subfigure}
    \hfill
    \begin{subfigure}[b]{0.32\textwidth}
        \centering
        \includegraphics[width=\textwidth]{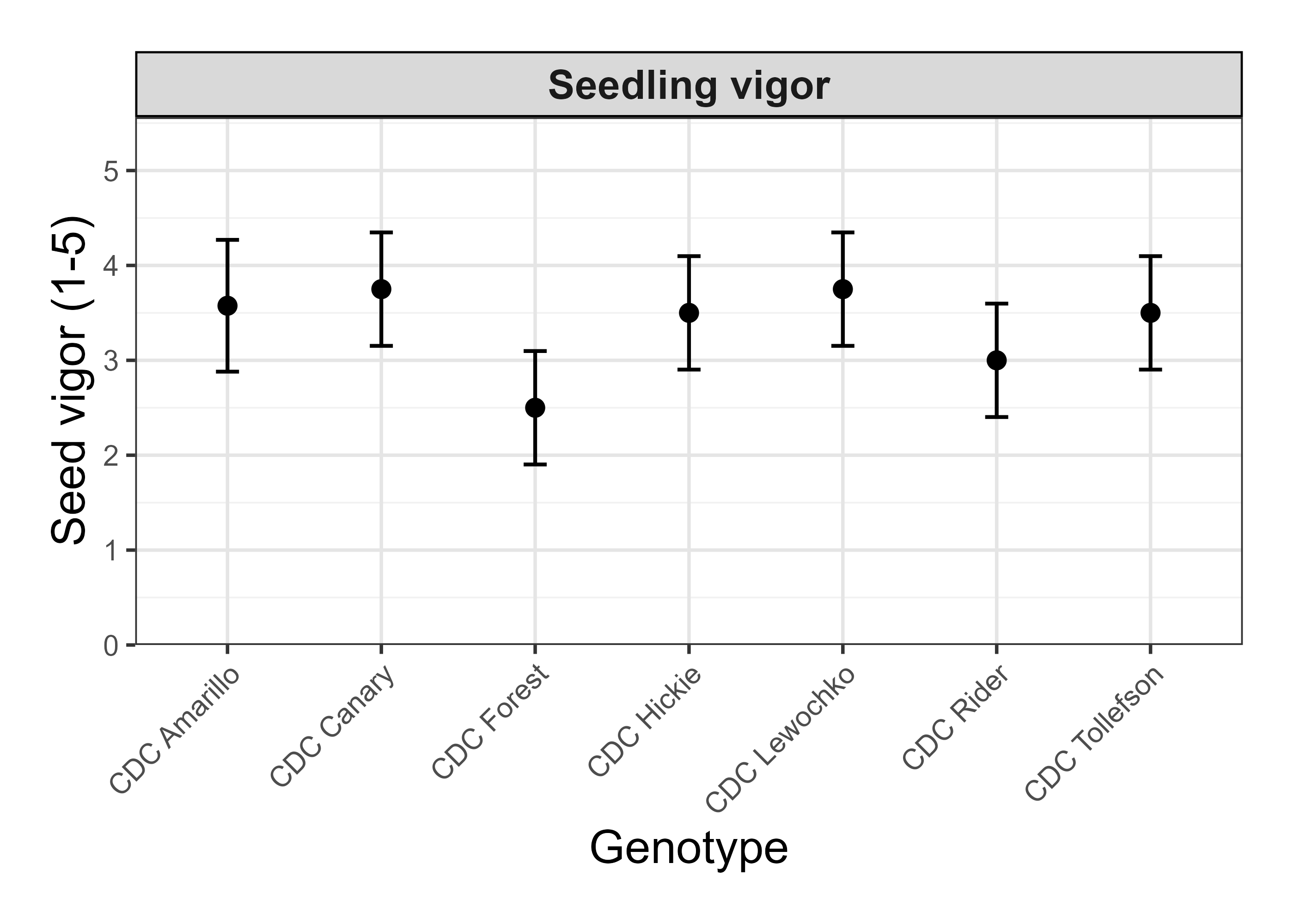}
        \caption{Seedling vigor}
        \label{fig:cad_sv}
    \end{subfigure}
    \hfill
    \begin{subfigure}[b]{0.32\textwidth}
        \centering
        \includegraphics[width=\textwidth]{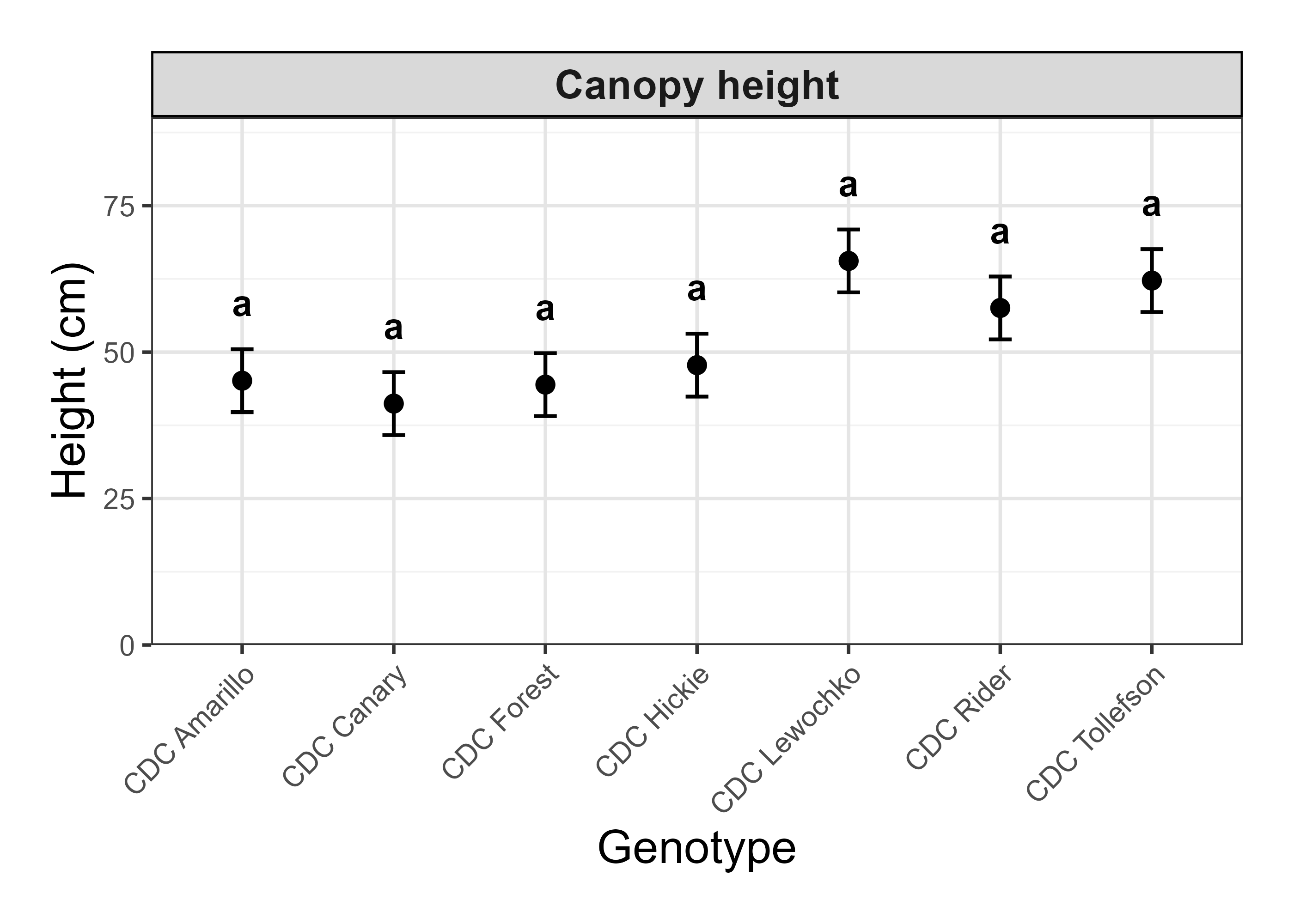}
        \caption{Canopy height}
        \label{fig:cad_canop_h}
    \end{subfigure}
    \vspace{0.3cm}
    \begin{subfigure}[b]{0.32\textwidth}
        \centering
        \includegraphics[width=\textwidth]{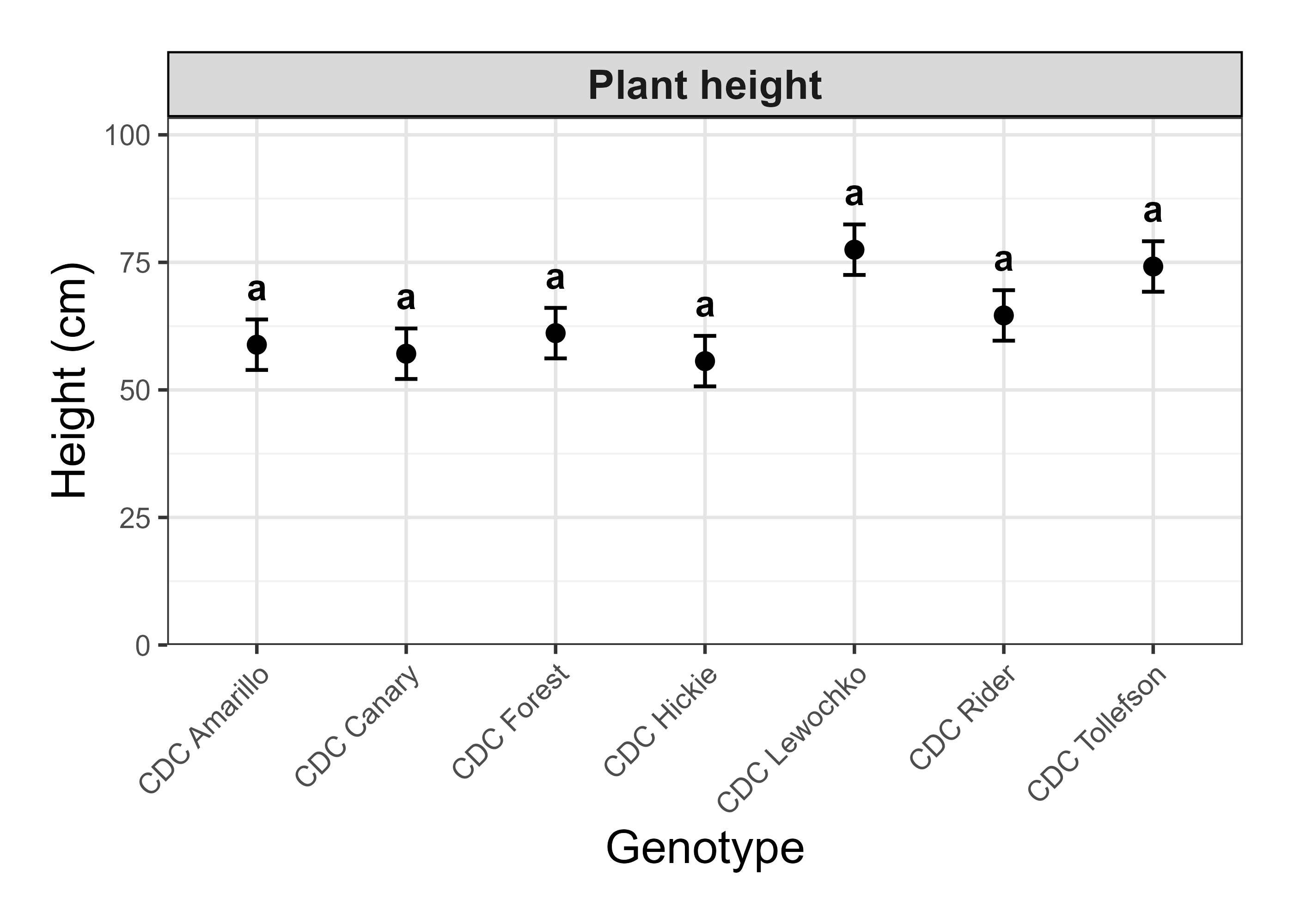}
        \caption{Plant height}
        \label{fig:cad_ph}
    \end{subfigure}
    \hfill
    \begin{subfigure}[b]{0.32\textwidth}
        \centering
        \includegraphics[width=\textwidth]{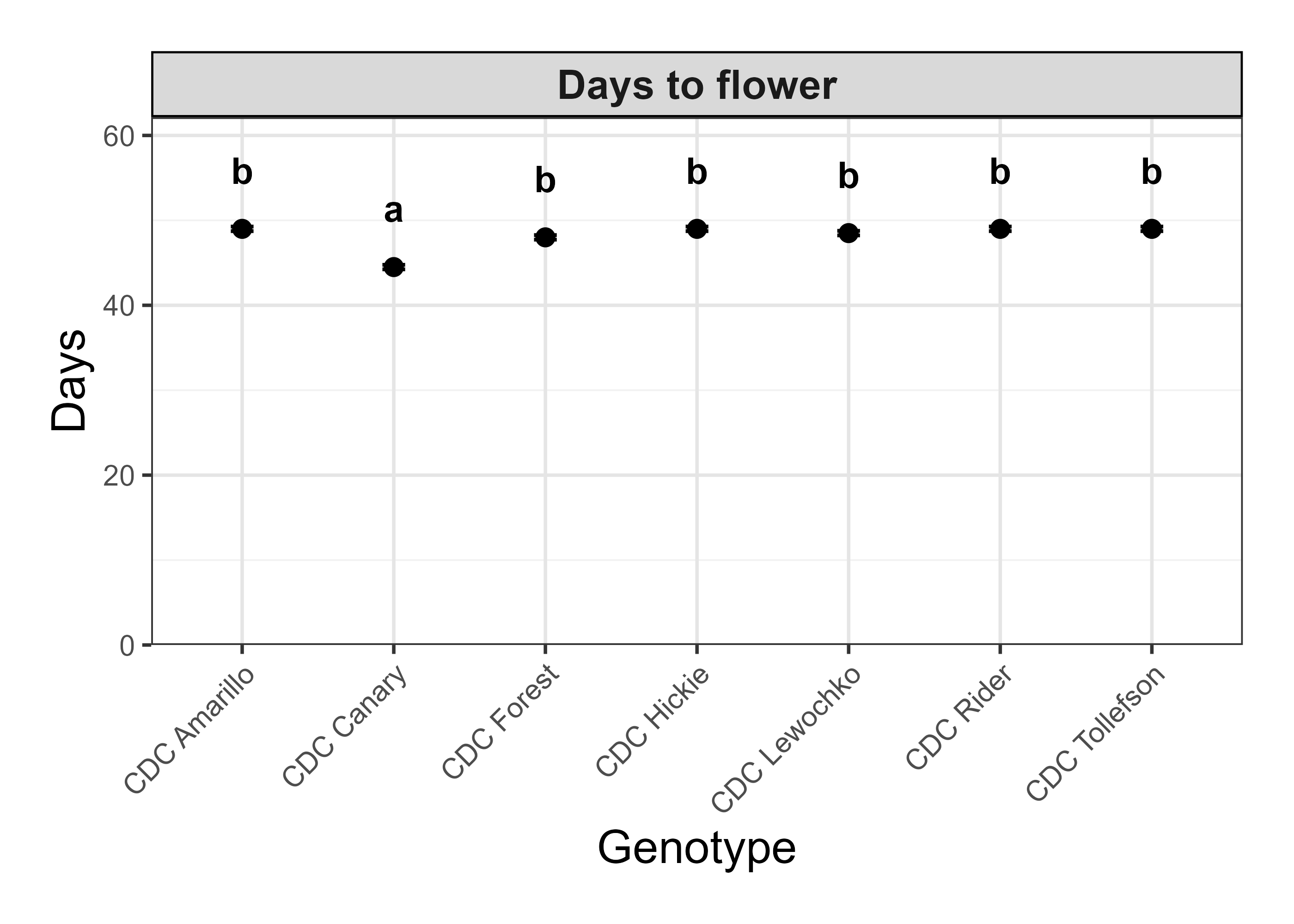}
        \caption{Days to flowering}
        \label{fig:cad_dtf}
    \end{subfigure}
    \hfill
    \begin{subfigure}[b]{0.32\textwidth}
        \centering
        \includegraphics[width=\textwidth]{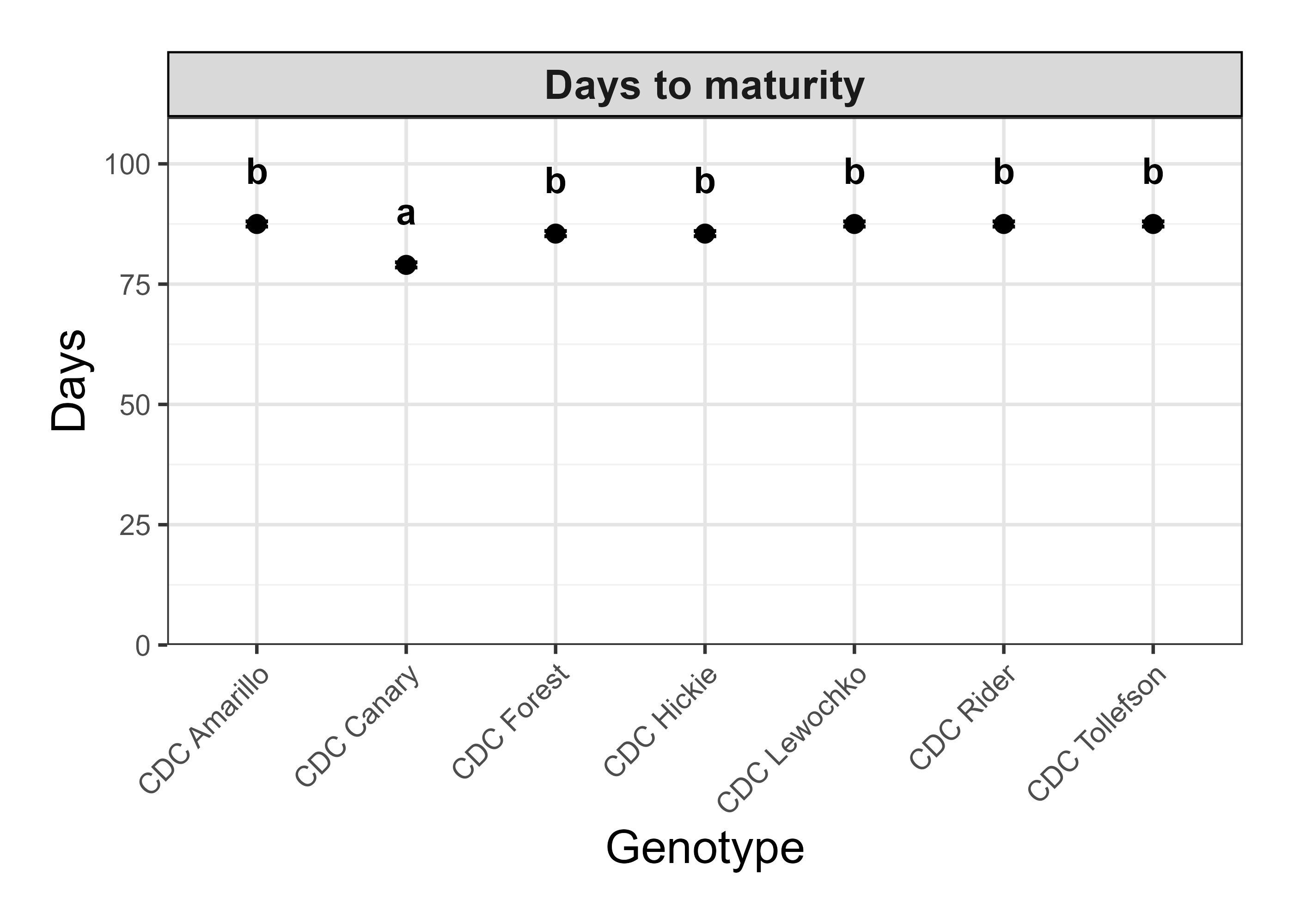}
        \caption{Days to maturity}
        \label{fig:cad_dtm}
    \end{subfigure}

    \caption{Means of yield and agronomic performance of seven Canadian genotypes across two site-years: (a) grain yield, (b) canopy height, (c) days to flowering (DTF), (d) days to maturity (DTM), (e) plant height, and (f) seedling vigor. The letter represents the significant difference between genotypes in the canopy height, plant height, days to flower, and days to maturity with Tukey's test.}
    \label{fig:cad_all_traits}
\end{figure}

\begin{figure}
    \centering
    \includegraphics[width=0.5\linewidth]{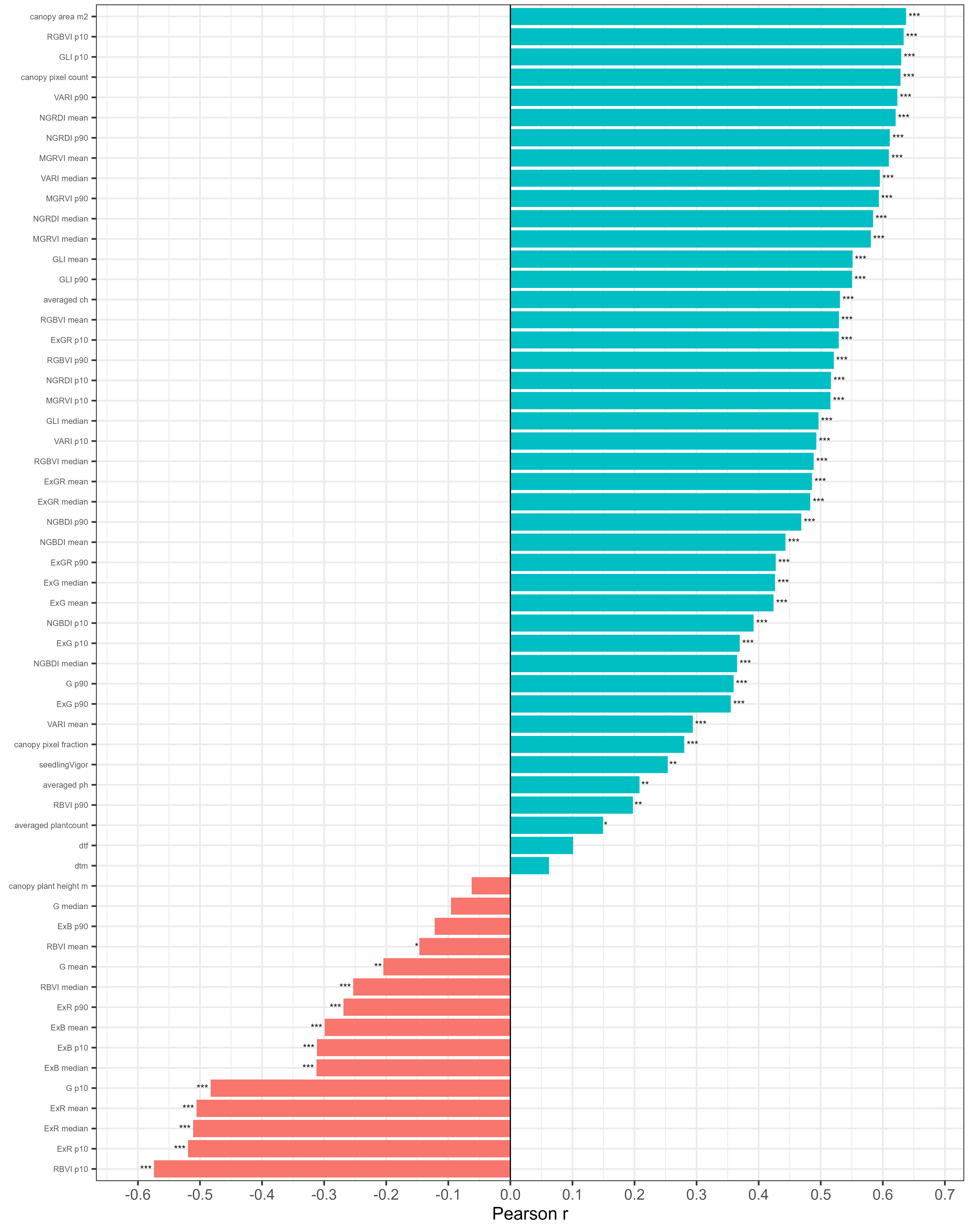}
    \caption{Pearson correlation of all UAV-derived and agronomic traits with seed yield.}
    \label{fig:supp_alltrait_yield}
\end{figure}

\clearpage
\section{Supplementary Tables}

\begin{table}[htbp]
\centering
\caption{RGB-based vegetation indices used in this study.}
\label{tab:rgb_indices}
\small
\begin{tabular}{p{6cm} p{2cm} p{5.2cm} p{1.0cm}}
\toprule
\textbf{Trait} & \textbf{Abbreviation} & \textbf{Formula} & \textbf{Citation} \\
\midrule

Green
& G
& $G/(G+R+B)$
& \\

Excess green index 
& ExG 
& $(2G - R - B)$ 
& \cite{woebbecke1995} \\

Excess blue index 
& ExB 
& $(1.4B - G)$ 
& \cite{meyer2008} \\

Excess red index 
& ExR 
& $(1.4R - G)$ 
& \cite{meyer2008} \\

Excess green-red index 
& ExGR 
& $\mathrm{ExG} - \mathrm{ExR}$ 
& \cite{meyer2008} \\

Normalized green-red difference index 
& NGRDI
& $\frac{G - R}{G + R}$ 
& \cite{gitelson2002} \\

Normalized green-blue difference index 
& NGBDI
& $\frac{G - B}{G + B}$ 
& \cite{bannari1995} \\

Modified green-red vegetation index 
& MGRVI 
& $\frac{G^{2} - R^{2}}{G^{2} + R^{2}}$ 
& \cite{bendig2015} \\

Red-Blue Vegetation Index
& RBVI
& $\frac{R^2 - B^2}{R^2 + B^2}$
& \cite{guo2020modified}\\

Visible atmospherically resistant index 
& VARI 
& $\frac{G - R}{G + R - B}$ 
& \cite{gitelson2002} \\

Green leaf index 
& GLI 
& $\frac{2G - R - B}{2G + R + B}$ 
& \cite{louhaichi2001} \\

Red-green-blue vegetation index 
& RGBVI
& $\frac{G^{2} - R \times B}{G^{2} + R \times B}$ 
& \cite{bendig2015} \\

\bottomrule
\end{tabular}
\end{table}

\begin{table*}[htbp]
\centering
\caption{Genetic variance components, heritability ($H^2$), and ANOVA statistics for agronomic and UAV-derived traits.}
\label{tab:supplementary_trait_stats}
\vskip 0.15in
\begin{small}
\resizebox{\textwidth}{!}{%
\begin{tabular}{lccccccccccccccc}
\hline
\textbf{Trait} & \textbf{$n\_F$} & \textbf{$F$-value} & \textbf{$df_{num}$} & \textbf{$df_{den}$} & \textbf{$p$-value} & \textbf{$H^2$} & \textbf{$V_g$} & \textbf{$V_{gxe}$} & \textbf{$V_e$} & \textbf{E} & \textbf{R} & \textbf{Mean $y$} & \textbf{$\sigma_e$} & \textbf{CV (\%)} \\ \hline
Seed yield & 204 & 3.60 & 16 & 80.00 & $6.90 \times 10^{-5}$ & 0.722 & 13918.49 & 24862.41 & 14475.64 & 6 & 2 & 468.37 & 120.31 & 25.69 \\
Canopy Height (manual) & 204 & 31.11 & 16 & 176.00 & $6.83 \times 10^{-43}$ & 0.968 & 133.10 & 0.00 & 53.05 & 6 & 2 & 40.19 & 7.28 & 18.12 \\ 
Plant Height (manual) & 204 & 17.72 & 16 & 80.00 & $5.80 \times 10^{-20}$ & 0.944 & 344.82 & 77.69 & 92.02 & 6 & 2 & 63.95 & 9.59 & 15.00 \\
Days to Flower & 204 & 54.67 & 16 & 80.00 & $2.94 \times 10^{-36}$ & 0.982 & 21.70 & 1.61 & 1.64 & 6 & 2 & 50.78 & 1.28 & 2.52 \\
Days to Maturity & 136 & 21.84 & 16 & 48.00 & $1.03 \times 10^{-16}$ & 0.954 & 44.64 & 7.01 & 3.11 & 4 & 2 & 85.17 & 1.76 & 2.07 \\
Seedling Vigor & 136 & 2.99 & 16 & 50.21 & 0.0016 & 0.666 & 0.35 & 0.53 & 0.33 & 4 & 2 & 3.11 & 0.58 & 18.51 \\
ExGR & 204 & 5.12 & 16 & 80.00 & $3.62 \times 10^{-7}$ & 0.805 & 0.0003 & 0.0003 & 0.0002 & 6 & 2 & -0.07 & 0.01 & 20.33 \\
AUC & 204 & 6.30 & 16 & 80.00 & $8.41 \times 10^{-9}$ & 0.841 & 249.23 & 179.75 & 205.17 & 6 & 2 & 99.40 & 14.32 & 14.41 \\
Canopy Area (mean) & 204 & 7.77 & 16 & 80.00 & $1.13 \times 10^{-10}$ & 0.871 & 0.16 & 0.08 & 0.11 & 6 & 2 & 2.49 & 0.34 & 13.46 \\
Peak Canopy (curve) & 204 & 10.59 & 16 & 80.00 & $7.94 \times 10^{-14}$ & 0.906 & 0.54 & 0.17 & 0.33 & 6 & 2 & 3.88 & 0.58 & 14.84 \\
DAP Max Growth (curve) & 204 & 4.56 & 16 & 80.00 & $2.37 \times 10^{-6}$ & 0.781 & 80.48 & 35.60 & 200.07 & 6 & 2 & 53.13 & 14.14 & 26.62 \\
Max Growth Rate (curve) & 204 & 2.51 & 16 & 80.00 & 0.0037 & 0.601 & 0.0003 & 0.0005 & 0.0013 & 6 & 2 & 0.10 & 0.04 & 35.70 \\ \hline
\end{tabular}%
}
\end{small}
\end{table*}

    